\documentclass[runningheads]{llncs}

\usepackage{eccv}

\usepackage{eccvabbrv}

\usepackage{booktabs}

\usepackage[accsupp]{axessibility}  

\usepackage{graphicx}
\usepackage{subcaption}
\usepackage{caption}
\usepackage{tabularx}
\usepackage{array}
\usepackage{xcolor}
\usepackage{wrapfig}

\usepackage{hyperref}

\usepackage{orcidlink}

\begin{document}

\title{POINav: Benchmarking and Enhancing Final-Meters Arrival in Real-World Vision-Language Navigation}


\author{Ruiyan Gong\inst{1,*} \and
Meisheng Zhang\inst{1,2,*,\ddag} \and
Yuxiang Zhao\inst{1,*} \and
Mingchao Sun\inst{1} \and
Yanfen Shen\inst{1} \and
Zedong Chu\inst{1} \and
Zhining Gu\inst{1} \and
Wei Guo\inst{1} \and
Xiaolong Cheng\inst{1} \and
Qiming Li\inst{1,\ddag} \and
Kangning Niu\inst{1} \and
Yanqing Zhu\inst{1} \and
Xiaolong Wu\inst{1} \and
Tianlun Li\inst{1,\dag} \and
Mu Xu\inst{1}}


\institute{Amap CV Lab, Alibaba Group \and
Peking University}

\maketitle
\begingroup
\renewcommand\thefootnote{*}
\footnotetext{Equal contribution.}
\renewcommand\thefootnote{\dag}
\footnotetext{Corresponding author.}
\renewcommand\thefootnote{\ddag}
\footnotetext{Work done during an internship at Amap CV Lab.}
\endgroup

\pagestyle{plain}

\begin{abstract}
    Real-world navigation is fundamentally driven by Points of Interest (POIs), yet reaching a precise POI remains a critical "final-meters" challenge. Existing Vision-Language Navigation (VLN) benchmarks of POI-goal navigation often suffer from coarse granularity or significant sim-to-real gaps due to generated scene. To bridge this gap, we present POINav-Bench, the first benchmark designed for closed-loop evaluation of real-world POI-goal navigation. It comprises 11 commercial areas reconstructed from real-world captures using 3D Gaussian Splatting (3DGS), covering 126,398 $m^{2}$ in total and spanning 163 distinct POIs. With traversability-aware annotations and reference trajectories, POINav-Bench enables high-fidelity evaluation of navigation agents in realistic, POI-rich real-world environments. Building on this, we propose the \textbf{POINav Brain-Action Framework} where a Brain module performs POI-grounded reasoning to guide an Action module in predicting continuous waypoints for real-world execution. We further curate the \textbf{POINav-Dataset}, containing 70K real-world signage-entrance pairs. Experiments show that our framework provides a viable path toward refining real-world POI-goal navigation.

\end{abstract}



\section{Introduction}
\label{sec:intro}

\begin{figure}[t]
    \centering
    \includegraphics[width=\linewidth]{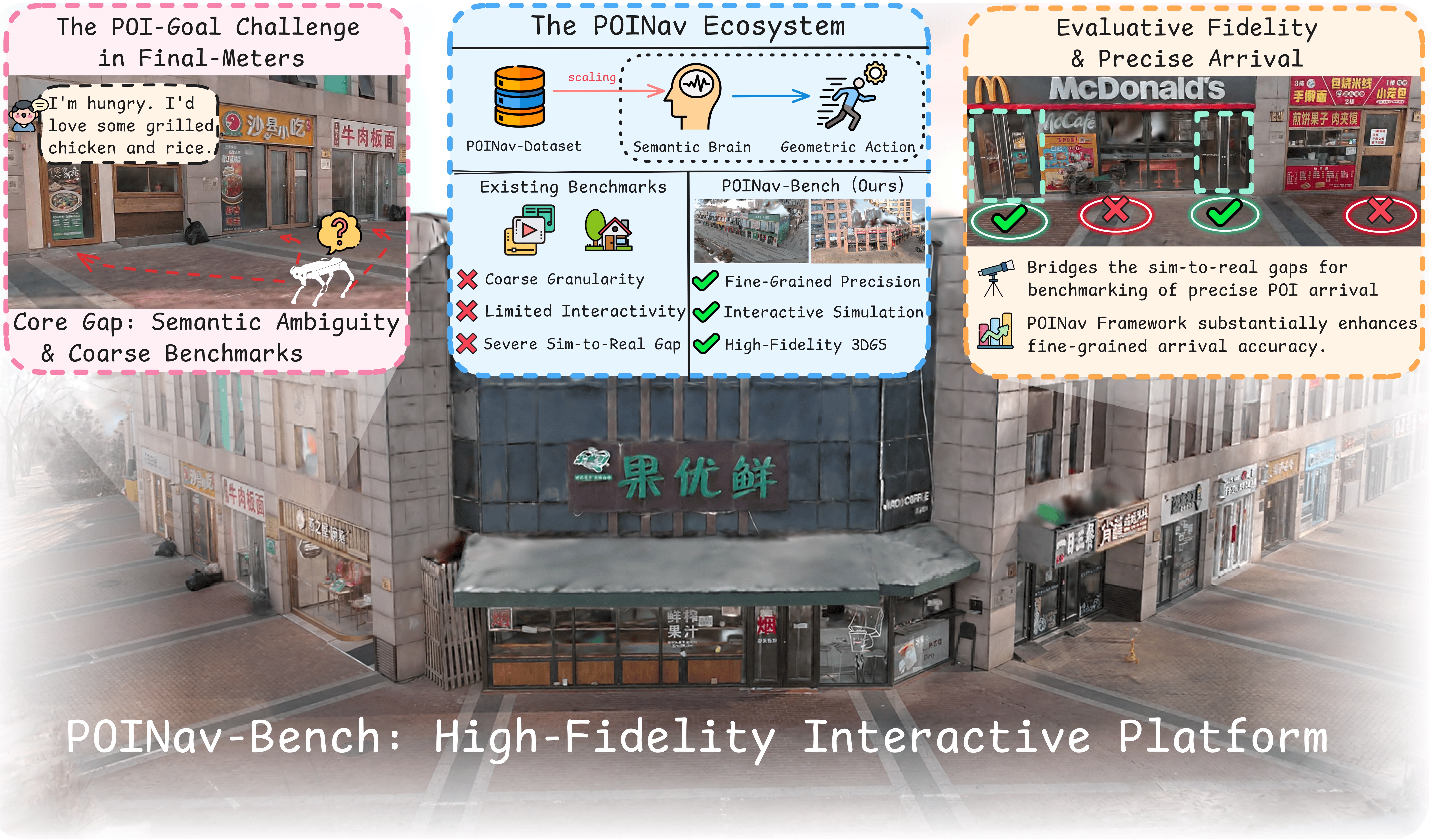} 
    \caption{\textbf{Overview of the POINav Ecosystem.} To overcome the limitations of semantic misalignments and coarse evaluation scales in final-meters scenarios, POINav-Bench establishes a high-fidelity, interactive 3DGS platform reconstructed from physical commercial streets. Furthermore, by coupling a newly curated grounding dataset with a decoupled brain-action architecture, this ecosystem empowers embodied agents to achieve physically precise POI arrivals and effectively close the sim-to-real gap.}
    \label{fig:page}
\end{figure}



The recent advancement of Vision-Language Models (VLMs) \cite{bai2025qwen3vltechnicalreport, guo2025seed15vltechnicalreport, wang2025internvl35advancingopensourcemultimodal, vteam2026glm45vglm41vthinkingversatilemultimodal} has established a powerful semantic foundation for embodied intelligence, unlocking unprecedented opportunities for Vision-Language Navigation (VLN) \cite{anderson2018vision}. Recent VLN research tends to undergo a notable paradigm shift, from theoretical simulations toward the challenges of real-world deployment, such as generating socially compliant trajectories that adhere to traffic norms and public conventions \cite{liu2025citywalker, chen2025socialnav}. Among VLN paradigms, point-goal navigation remains the most fundamental task; however, real-world navigation is inherently POI-driven, as Points of Interests (POIs), such as retail shops, serve as the primary origins and destinations that anchor human activities in physical environments. We posit that reconciling geometric goal specification with semantically grounded POI understanding is fundamental to deploying VLN agents effectively in real-world, human-centric environments.

POI-goal navigation \cite{zhao2026bridgingindooroutdoorgapvisioncentric, chu2026abot} targets the final navigation phase, requiring an embodied agent to navigate to the precise access point of a specific POI, defined as a discrete geographical entity with a specific functional identity (e.g., a retail shop or service counter), after arriving at its vicinity. Crucially, this must be achieved without relying on external priors such as high-precision coordinates \cite{wang2024interactive} or semantic maps \cite{chaplot2020object}. 

However, the community still lacks a closed-loop benchmark for POI-goal navigation in real-world settings. Existing benchmarks suffer from critical limitations: some rely on non-panoramic or synthetic trajectory data \cite{zhao2026bridgingindooroutdoorgapvisioncentric}, inducing sim-to-real gaps; others, such as CitySeeker \cite{wang2025cityseeker}, operate at a coarse street-segment level that misaligns with the fine-grained precision requirements of POI-goal navigation. Consequently, there is an urgent need for a fine-grained, up-to-date, and simulation-compatible benchmark that bridges this gap and empowers precise final-meters navigation in real-world scenarios.

To address this, we introduce \textbf{POINav-Bench}, a high-fidelity evaluation benchmark centered on 11 distinct large-scale commercial areas with high POI density. Leveraging high-precision LiDAR and photogrammetry, we conducted extensive on-site data collection after February 2026, reconstructing these freshly captured real-world environments into high-quality 3D Gaussian Splatting (3DGS) \cite{kerbl20233d} scenes, encompassing hundreds of realistic POIs. Crucially, we integrate these assets into the Isaac Sim platform \cite{NVIDIA_Isaac_Sim}. While specifically designed to provide a rigorous benchmark for final-meters POI-goal navigation, the photorealistic and geometrically accurate nature of our 3DGS assets offers a versatile foundation for developing benchmarks across diverse vision and robotics tasks. To maximize its impact and facilitate reproducibility, we will open-source the complete 3DGS assets for the broader research community.

Furthermore, to refine the paradigm of POI-goal navigation, we propose \textbf{POINav}, a novel VLN framework, which decouples the task into two stages: POI-grounded Reasoning (the brain module) and Global-Context Action Querying (the action module). Our brain module employs a consecutive spatially conditioning mechanism to first localize POI signage as a semantic anchor, which then guides the regression of the target entrance as the final affordance. To support this POI-grounded reasoning, we introduce \textbf{POINav-Dataset}, a large-scale dataset comprising 70K real-world samples. For the action module, we simplify the architecture of BridgeNav \cite{zhao2026bridgingindooroutdoorgapvisioncentric} by excluding several task-specific modules.

Our main contributions can be summarised as:

\begin{itemize}
    \item \textbf{POINav-Bench}: We introduce the first high-fidelity benchmark for fine-grained POI-goal navigation, featuring 11 large-scale commercial areas reconstructed via 3D Gaussian Splatting (3DGS) from fresh, post-2026 LiDAR and photogrammetry data. Integrated into Isaac Sim, it bridges the sim-to-real gap and provides a rigorous, physically accurate testbed for final-meters navigation that existing coarse or synthetic benchmarks lack.

    \item \textbf{POINav Framework \& Dataset}: We propose POINav, a novel VLN paradigm that decouples navigation into POI-grounded reasoning and Global-Context Action Querying. To enable this, we curate POINav-Dataset, comprising 70K real-world samples that facilitate robust POI-grounded reasoning without external priors.

    \item \textbf{Open-Source POINav Ecosystem}: To accelerate advancements in precise real-world navigation, we plan to open-source the full POINav-Bench and strive to make the POINav-Dataset publicly available, creating a robust ecosystem for vision-centric navigation and general robotics research.
\end{itemize}

\section{Related Work}
\subsection{Benchmarks for Goal-Oriented Navigation}





Goal-oriented navigation \cite{ieong2025multimodal} has evolved through several key paradigms, including point-goal \cite{wijmans2019dd, zhao2021surprising}, object-goal \cite{batra2020objectnav, yokoyama2024vlfm, zhang2024uni}, image-goal \cite{anderson2018vision, bono2023end}, and instruction-based navigation \cite{cheng2024navila, long2024instructnav, zhang2025embodied, wei2025streamvln}. These approaches are evaluated via specialized benchmarks: while point-goal tasks prioritize coordinate-based accuracy and social adherence \cite{liu2025citywalker, chen2025socialnav}, object-goal \cite{yokoyama2024hm3d} and image-goal benchmarks \cite{krantz2022instance} challenge agents to navigate using semantic categories or visual references, respectively. Furthermore, instruction-based benchmarks \cite{krantz2020beyond, ku2020room} shifts the focus toward semantic grounding, covering the spectrum from basic instructions to intricate, long-horizon missions.




    

POI-goal navigation exhibits similarities with these paradigms; however, POI-related benchmarks for navigation remain notably limited. The recent CitySeeker \cite{wang2025cityseeker} represents a notable yet limited attempt; it operates at a coarse street-segment level, failing to provide the fine-grained, meter-scale action space essential for embodied robots in precise final-meters scenarios. Additional, BridgeNav \cite{zhao2026bridgingindooroutdoorgapvisioncentric} evaluates navigation through an open-loop protocol, where models are assessed by single-step waypoint prediction success rates under 0.1–0.3 m thresholds and navigation efficiency. Nevertheless, because the trajectories are sampled from AI-generated videos, its evaluation primarily measures distributional alignment with the generated trajectory data. As a result, a high BridgeNav score may indicate good imitation of the training distribution, but it cannot faithfully characterize a model’s ability to interact with an environment and perform real closed-loop point-to-point navigation.


This limitation highlights the urgent need for a fine-grained, closed-loop benchmark that supports POI-goal navigation.

\subsection{Paradigm Shift and POI-goal Navigation}

A common end-to-end paradigm \cite{dorbala2022clip, zheng2024towards} formulates navigation actions as textual tokens and treats the task as next-token prediction within VLMs, enabling VLMs to process multimodal instructions and autoregressively decode low-level actions. However, this approach suffers from significant limitations, most notably error compounding in the discretized action space induced by LLM decoding. Recent VLN approaches start to shift from direct action prediction to a dual-system design \cite{chen2025socialnav, wei2025ground, xue2025omninav} that decouples semantic understanding and action execution. Typically, a VLM is designed to be dedicated to high-level semantics understanding, while another module generates actions based on the VLM’s output through a non-autoregressive way.

BridgeNav \cite{zhao2026bridgingindooroutdoorgapvisioncentric} builds upon a foundational Vision-Language model, Qwen2.5-VL-3B \cite{bai2025qwen25vltechnicalreport}, but significantly extends it by integrating several additional task-specific modules. This complex integration significantly complicates the training process. BridgeNav is trained in two stages: the first stage focuses exclusively on optimizing the bounding box head, while the second stage involves multi-objective optimization, which likely requires a large amount of data to converge. Moreover, due to its specialized architecture and training scheme, BridgeNav is inherently difficult to integrate with the prevailing dual-system navigation paradigm, making it challenging to build a unified system capable of handling multiple types of navigation within a single framework.

Hence, we present a novel dual-system approach for POI-goal navigation, paving the way for a unified embodied navigation system that can integrate POI-goal paradigm with multiple goal paradigms.

\section{POINav-Bench}


We present POINav-Bench to evaluate final-meters navigation of embodied agents in visually ambiguous, POI-dense environments. The benchmark comprises eleven 3D Gaussian Splatting (3DGS) scenes of commercial areas, reconstructed using high-precision LiDAR and photogrammetry. To mitigate privacy concerns, all pedestrians have been removed from the 3D reconstructions. Additionally, every POI in these scenes was manually annotated. POINav-Bench involves 163 distinct POIs and spans 126,398 $m^2$. Finaly, all evaluations are simulated on NVIDIA Isaac Sim \cite{NVIDIA_Isaac_Sim}.

\subsection{Problem Formulation}

\paragraph{\textbf{Task Definition.}}

We formulate the POI-goal navigation task as a finite-horizon Partially Observable Markov Decision Process (POMDP) situated in a continuous, physically grounded simulator. At the beginning of an episode ($t=0$), the embodied agent receives a natural language instruction $L$ specifying a target POI $g$, and its initial pose $p_0$ is sampled from a traversability-aware spatial distribution. At each timestep $t \in \{0, \dots, T\}$, the agent captures an egocentric visual observation $V_t$ and executes a continuous geometric action $a_t \in \mathcal{A}$, parameterized as local waypoint increments $\Delta p$. This action sequentially updates the agent's pose via the underlying environmental transition dynamics $\mathcal{P}$, yielding a subsequent observation governed by the emission model $\mathcal{O}$:

\begin{equation}
\label{eq:pomdp_transition}
p_{t+1} \sim \mathcal{P}(\cdot \mid p_t, a_t), \quad V_{t+1} \sim \mathcal{O}(\cdot \mid p_{t+1})
\end{equation}

The agent's objective is to learn a multimodal navigation policy $\pi$ that maps the semantic instruction $L$ and the accumulated observation history $\mathcal{H}_t = \{V_0, \dots, V_t\}$ to the action space:

\begin{equation}
\label{eq:policy}
a_t \sim \pi(\cdot \mid \mathcal{H}_t, L)
\end{equation}

Unlike abstract point-goal navigation, each POI target $g$ in POINav-Bench is structurally bound to a ground-truth physical entrance region $\mathcal{E}_g$, defined as a horizontal ground-flush bounding box. The task requires the agent to physically arrive at this semantic target; an episode is deemed successful if and only if the agent's terminal pose $p_T$ satisfies $d(p_T, \mathcal{E}_g) \leq \epsilon$, where $d(\cdot)$ denotes the spatial distance to the horizontal region boundaries and $\epsilon$ is a strict proximity threshold, provided that the entire trajectory remains collision-free across all traversable surfaces.


\paragraph{\textbf{Success criterion.}} We define a strict success criterion for POINav: a navigation episode is considered successful only if the agent reaches the vicinity of the target POI’s entrance, within a pre-defined distance threshold, without colliding with any obstacles during simulation.

\subsection{POINav-Bench Construction}
\begin{figure}[htbp]
    \centering
    \includegraphics[width=\textwidth]{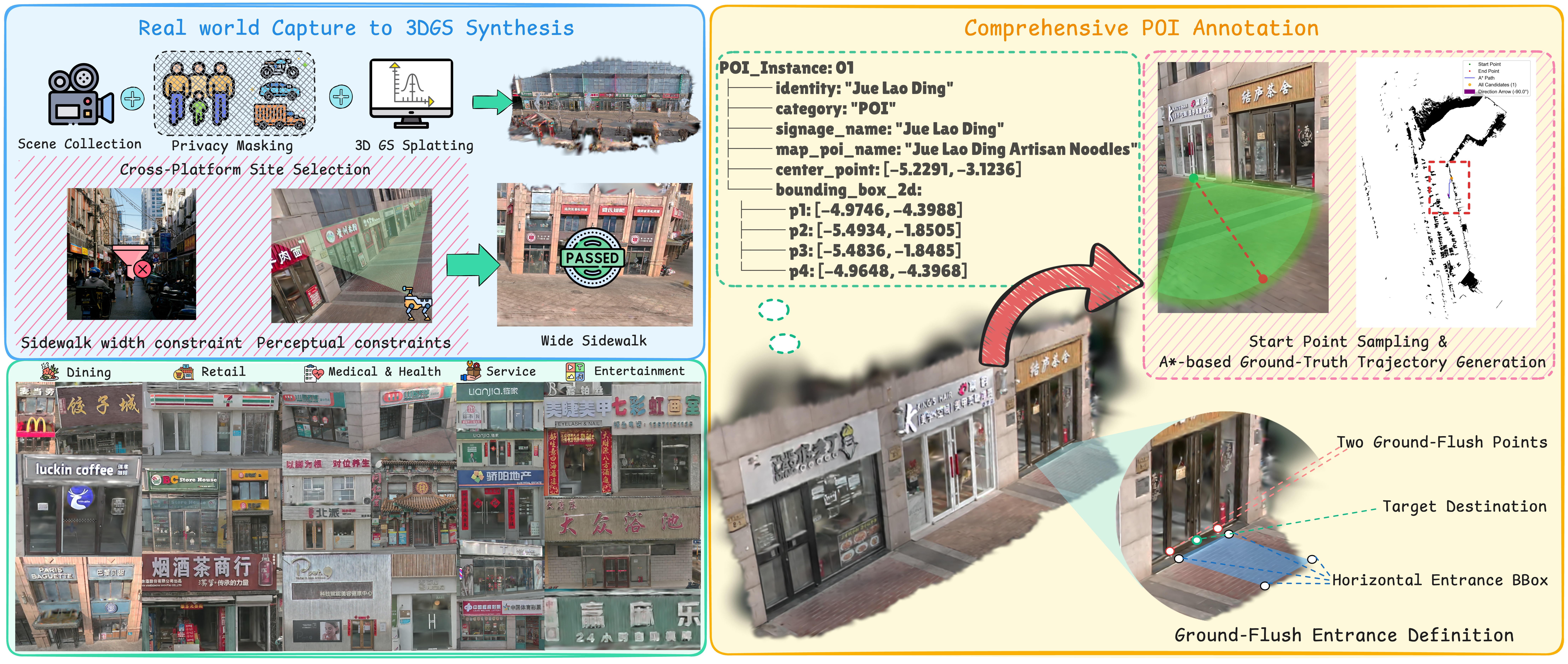}
    \caption{\textbf{POINav-Bench construction pipeline.} High-fidelity scenes are first reconstructed using 3DGS captures with privacy-masked mechanism, with sites selected for perceptual consistency (\textbf{Top Left}). Diverse POIs are then identified across multiple categories \textbf{(Bottom left}) and annotated with rich semantic metadata and spatial anchors (\textbf{Right}). Specifically, entrances are annotated by horizontal bounding boxes extending between two flush points to cover their full width. Start point sampling and A*-based ground-truth trajectories support to evaluate the navigation efficiency.}
    \label{fig:benchmark}
\end{figure}

\paragraph{\textbf{Site Selection and Scene Acquisition.}} We selected 11 diverse commercial areas characterized by high POI density and a wide variety of POI types. A critical criterion for site selection was sufficient sidewalk width. Specifically, quadruped robots operate with a low camera mounting height and possess a limited vertical field of view; in narrow streets, such physical constraints can block the robot’s line of sight to POI signage. Following site confirmation, we conducted comprehensive data capture at each location to obtain high-fidelity 3DGS scenes.

\paragraph{\textbf{3DGS Scene Annotation.}} We manually annotate POIs by defining entrances through horizontal bounding boxes derived from dual flush points. These boxes serve as ground-truth regions for evaluating agent arrival accuracy. Each POI is assigned its respective attributes, and any instances with unclear signage are omitted from the dataset.



\paragraph{\textbf{Start Point Sampling.}} For each annotated POI in POINav-Bench, we generate multiple start positions by randomly sampling within a pre-defined radius, constrained to valid traversable areas. Specifically, we manually define walkable regions on the 3DGS reconstruction of each scene using irregular road networks that exclude non-walkable surfaces (e.g., grass) and impassable obstacles such as large steps unsuitable for quadrupedal robots. This ensures all sampled start points lie on feasible terrain. Finally, we manually validate each start location to ensure the agent, from a quadruped-level viewpoint, can perceive at least partial visual cues of the signage in terms of the target POI.

\paragraph{\textbf{Ground-Truth Trajectory Annotation.}} To enable evaluation of navigation efficiency, we generate ground-truth trajectories from each start position to the target POI. Leveraging  MoGe-V2 \cite{wang2025moge}, we first segment navigable regions in the 3D scene by analyzing surface normals: points with upward-facing normals are classified as traversable ground, while others (e.g., walls, poles) are treated as obstacles. This segmentation is projected onto a 2D top-down plane to produce an agent-centric local occupancy grid, on which we run A* algorithm to compute the shortest feasible path, serving as the ground-truth trajectory.

\paragraph{\textbf{Instruction Setup.}} Given the architectural diversity of current VLN systems, we avoid prescribing a fixed instruction template or prompt in POINav-Bench for evaluating goal-reaching performance. Instead, we provide only the annotated POI name as the high-level goal, enabling seamless integration of a wide range of VLN approaches. This design choice reflects the real-world complexity of POI naming: POI names are not always derived directly from visible signage text but may stem from logos, local conventions, or map-based metadata, making them inherently ambiguous and context-dependent. By supplying the canonical POI name, our benchmark remains flexible while preserving the core challenge of grounding language to physical locations.






\section{Methodology}

We present the POINav Brain-Action Framework, which bridges abstract semantic instructions and precise geometric execution for final-meters navigation. Our approach decomposes the task into two complementary modules. The Semantic Brain Module acts as a high-level spatial planner: instead of reducing targets to discrete pixel goals \cite{cheng2024navila, xue2025omninav, wei2025ground, huang2026ticvlathinkincontrolvisionlanguageactionmodel}, it grounds the target into hierarchical visual references that preserve its full spatial extent. The Geometric Action Module, in contrast, serves as a reactive motor policy that aggregates temporal context from the observation history to generate continuous trajectory waypoints. These two modules are coupled via spatial conditioning: the geometric boundaries produced by the Brain Module are encoded as visual cues to guide the action policy.


\subsection{Semantic-to-Geometric POI Grounding}

\paragraph{\textbf{From Pixel Goals to Hierarchical Visual References.}}
Prevailing vision-and-language navigation paradigms increasingly rely on VLMs \cite{zhang2025embodied}, yet they typically reduce semantic targets to discrete pixel goals. This compression discards the spatial extent of the navigation target: a complex physical entrance is collapsed to a single coordinate. Without explicit spatial boundaries, planning modules fail to establish reliable visual references and cannot identify the correct physical stopping condition \cite{lin2025vlnversebenchmarkvisionlanguagenavigation}. In practice, a deviation of just a few pixels can shift the target from a navigable door to an impassable glass wall, causing unstable approach trajectories and incorrect orientations.

We argue that the provided visual cues, which guide the action module, require preserving the full spatial extent of the goal through a region-based constraint. The Semantic Brain Module grounds the target as a complete physical entity rather than a point coordinate. This regional representation provides the Geometric Action Module with tractable visual references, stabilizing the approach trajectory and ensuring precise arrival at the physical entrance.

\paragraph{\textbf{POI-Grounded Reasoning.}}
We cast the POI grounding problem as a hierarchical dependency Reasoning task. Let $L$ denote the natural language instruction and $V_t$ represent the egocentric observation at timestep $t$. We define the visual goal $\mathcal{G}_t$ as a entity set consisting of a POI signage $b_{id}$ and a physical entrance $b_{geo}$, where $b \in \mathbb{R}^4$ represents the bounding box coordinates. The objective of the Brain Module is to maximize the likelihood of localizing the entry point conditioned on the target identity:

\begin{equation}
\label{eq:hierarchical_grounding}
P(\mathcal{G}_t | V_t, L) \propto \underbrace{P(b_{id} | V_t, L)}_{\text{Semantic Anchoring}} \cdot \underbrace{P(b_{geo} | b_{id}, V_t, L)}_{\text{Geometric Localization}}
\end{equation}

The first term $P(b_{id} | V_t, L)$ governs the \textbf{Semantic Anchoring} of the target, acquiring the global POI reference (e.g., a high-mounted shopboard) from a distance to maintain robustness against urban distractors. The second term $P(b_{geo} | b_{id}, V_t, L)$ performs \textbf{Geometric Localization} of the physical entrance, conditioned explicitly on the acquired signage. We constrain the probability of the entrance to depend on the signage, establishing a structural spatial prior that forces the system to localize the physical entry point directly from the established semantic context, effectively preventing the mislocalization of entrances on unrelated POIs.

\begin{figure}[t]
    \centering
    \includegraphics[width=\linewidth]{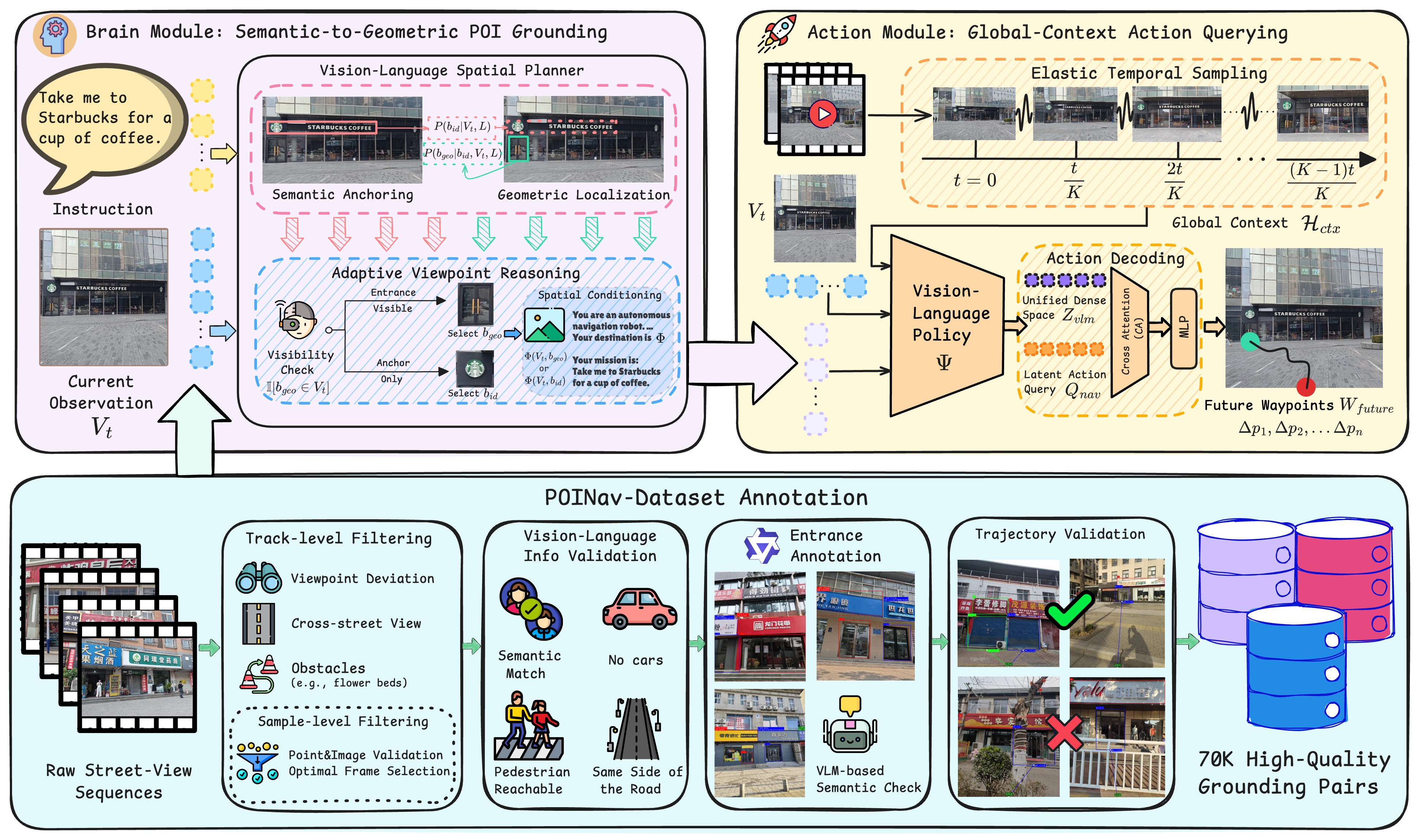} 
    \caption{Overview of the POINav Framework. (Top) The \textbf{Semantic Brain Module} grounds POI targets into explicit visual references. Conditioned on these visual cues, the \textbf{Geometric Action Module} predicts continuous trajectory waypoints. (Bottom) The automated \textbf{Dataset Pipeline} curates high-quality signage-entrance pairs for grounding supervision.}
    \label{fig:method}
\end{figure}

\paragraph{\textbf{Data Pipeline for POI-grounded Reasoning.}}
Continuous trajectory data is inherently scarce, and current diffusion-based generation methods often produce physically inconsistent results, limiting the direct scaling of motor policies \cite{zhao2026bridgingindooroutdoorgapvisioncentric, wang2025cityseeker, chu2026abot}. Rather than scaling trajectory collection, we scale at the semantic grounding level: by improving the spatial reasoning of the Brain Module, we enhance POI navigation without requiring massive trajectory datasets. To this end, we construct the POINav Grounding Dataset using a multi-stage automated pipeline over street-view sequences captured in real-world urban environments, curating explicit signage–entrance pairs with strict semantic and physical consistency.

The pipeline begins with track and sample-level filtering to discard frames with extreme camera rotations, cross-street views, or severe occlusions, retaining only observations with verified semantic alignment to the target POI. Next, we impose geometric constraints to localize the physical entrance. The vision-language foundation model predicts the entrance bounding box conditioned on the detected POI signage. Subsequent rule-based validation actively filters out physically implausible annotations, such as floating boxes or spurious predictions on solid walls.

Finally, we perform a physical traversability check. A straight-line trajectory from the camera origin to the predicted entrance is validated to ensure it lies on a horizontal plane, remains free of obstacles, and does not originate off-road or require street crossings. This curation process yields 70k high-quality training pairs, providing the scalable supervision necessary to learn the conditional mapping from POI signages to physical entrances. These hierarchical visual references empower the Brain Module to significantly reduce the spatial decision-making burden on the downstream continuous control policy.




\paragraph{\textbf{Adaptive Viewpoint Reasoning.}}
A critical advantage of this decoupled design is its robustness to partial observability and extreme viewpoint variations during quadrupedal locomotion. In urban environments, the visual availability of the POI signage and the physical entrance evolves sequentially. At a distance, the high-mounted POI signage is highly salient while the entrance appears geometrically coarse. Conversely, during physical arrival, the robot's low sensor profile often causes the signage to exit the vertical field of view, leaving only the entrance visible. To address this, we define the dynamic visual cue $C_t$ provided to the Action Module as a visibility-conditioned image cropping mechanism.

\begin{equation}
\label{eq:adaptive_prompt}
C_t = \mathbb{I}[b_{geo} \in V_t] \cdot \Phi(V_t, b_{geo}) + (1 - \mathbb{I}[b_{geo} \in V_t]) \cdot \Phi(V_t, b_{id})
\end{equation}

where $\Phi(V_t, b)$ denotes the cropping operation that extracts the localized image patch from the current observation $V_t$ using the predicted bounding box $b$, and $\mathbb{I}[\cdot]$ is the visibility indicator function. This mechanism executes a deterministic visual handover as the agent approaches the target. When the physical entrance $b_{geo}$ becomes visually resolvable, the system crops this region to serve as the dominant visual reference for precise alignment. Otherwise, it falls back to extracting the POI signage $b_{id}$ to provide a salient directional bearing. By transmitting explicit image patches, we formulate the downstream execution as a robust visual reference navigation task, significantly reducing the spatial reasoning burden on the continuous action policy.

\subsection{Global-Context Action Querying}

The Action Module performs continuous geometric planning by combining the spatial constraints from the Brain Module with temporal cues from the observation history. Unlike the Brain Module, which operates on individual snapshots, the Action Module must infer velocity, acceleration, and approach progress from the observation stream. We formalize this via two mechanisms: Elastic Temporal Sampling for global context retention and Latent Action Querying for efficient geometric decoding.

\paragraph{\textbf{Elastic Temporal Sampling.}}
Standard sliding-window approaches in visual navigation constrain the agent's memory to only its recent observation history. During physical arrival, however, generating accurate waypoints requires maintaining continuous spatial awareness of the entire approach trajectory. We mitigate this myopia via \textbf{Elastic Temporal Sampling}, which maintains a compressed yet globally coherent visual memory. Let $\mathcal{H}_{raw} = \{V_0, ..., V_t\}$ denote the full episode history. Rather than discarding older frames, we construct a sparse context set $\mathcal{H}_{ctx}$ by sampling $K$ frames based on a time-dilated index function:

\begin{equation}
\label{eq:elastic_sampling}
\mathcal{H}_{ctx} = \{V_{\tau_k} \mid \tau_k = \lfloor k \cdot \frac{t}{K} \rfloor, k \in \{0, ..., K\}\} \cup \{V_t\}
\end{equation}

This uniform sampling strategy ensures that the temporal coverage spans the entire episode duration $[0, t]$ while maintaining constant computational complexity. As the episode progresses, the sampling interval adaptively dilates, preserving the global motion trend and relative progress cues required to regulate approach velocity without overwhelming the context window of the visual backbone.

\paragraph{\textbf{Latent Action Querying.}}
To convert these high-dimensional visual representations into precise geometric actions, we depart from the computationally intensive iterative diffusion processes common in recent continuous control policies \cite{xue2025omninav,wei2025ground,chen2025socialnav,chu2026abot}. Instead, we frame continuous control as a deterministic latent querying process, repurposing a pretrained vision-language backbone to bypass the latency of iterative sampling. The backbone of the Action Module processes the sampled history $\mathcal{H}_{ctx}$ and the current observation $V_t$. Crucially, the model is directly conditioned on the dynamic visual cue $C_t$ derived in Eq. \ref{eq:adaptive_prompt}, effectively guiding the backbone's attention toward the explicit visual reference rather than abstract coordinates.

\begin{equation}
\label{eq:visual_encoding}
Z_{vlm} = \Psi(\mathcal{H}_{ctx} \cup \{V_t\}, C_t)
\end{equation}

where $\Psi$ represents the autoregressive multimodal backbone mapping the interleaved spatio-temporal visual tokens and the spatial text prompt into a unified dense latent space. To extract navigation-critical geometry from these multimodal latents $Z_{vlm}$, we introduce a set of learnable latent queries $Q_{nav} \in \mathbb{R}^{N \times d}$. These queries interact with the contextualized spatio-temporal latents via a Cross-Attention (CA) mechanism acting as an information bottleneck:

\begin{equation}
\label{eq:action_decoding}
Z_{geo} = \text{CrossAttn}(Q_{nav}, Z_{vlm}), \quad \mathcal{W}_{future} = \text{MLP}(Z_{geo})
\end{equation}

The resulting geometric latents $Z_{geo}$ capture the essential spatial structures of the target affordance. Finally, a lightweight Multi-Layer Perceptron decodes these latents into a sequence of waypoints $\mathcal{W}_{future} = \{\Delta p_1, ..., \Delta p_H\}$, enabling the agent to execute smooth, continuous trajectory tracking aligned precisely with the grounding predictions of the Brain Module.




\section{Solution to POINav-Bench}
\subsection{Experimental Results}

\begin{table}[tbp]
\small
\centering
\renewcommand{\arraystretch}{1.2}
\setlength{\tabcolsep}{8pt}
\begin{tabular}{lcc}
\toprule
\textbf{Method} & \textbf{SR (2m) $\uparrow$} & \textbf{SPL $\uparrow$} \\
\midrule
ViNT \cite{shah2023vint}              & 19.01 & 18.24 \\
OmniNav (Vanilla)$^\star$ & 23.92 & 22.44 \\
OmniNav \cite{xue2025omninav}           & 34.36 & 31.52 \\
POINav (Ours)            & \textbf{42.33} & \textbf{40.29} \\
\bottomrule
\end{tabular}
\caption{\textbf{Navigation Evaluation.} All methods use the same backbone: Qwen2.5-VL-3B. $^\star$ denotes the model is \textbf{not trained} on the BridgeNav dataset; all other methods are trained on BridgeNav dataset.}
\label{navigation_metric}
\end{table}


Most existing VLN models are trained on datasets without similar navigation scenarios, leading to weak and meaningless performance on our benchmark. We do not intend to retrain all mainstream VLN architectures for architectural comparison.


We train three representative VLN frameworks, namely ViNT~\cite{shah2023vint}, OmniNav, and our POINav, on the BridgeNav dataset using Qwen2.5-VL-3B as the shared backbone, and evaluate them on POINav-Bench. Since vanilla OmniNav possesses inherent OCR capabilities relevant to POI signage recognition, we additionally include it as an untrained baseline. Table~\ref{navigation_metric} summarizes the quantitative results. In the experiments, the initial position of the embodied agent is set outside a 10m radius from the target POI. We report Success Rate and Success Rate (2m), which measure the proportion of episodes where the agent precisely reaches the target POI and where the final position lies within 2m linear distance of the target location, respectively. As shown in the table, POINav, by leveraging cooperative reasoning between the fast and slow systems, outperforms the competing method in navigation success rate. In addition, we report Success Path Length (SPL) to further evaluate the navigation efficiency of POINav.

\begin{figure}[t!]
    \centering
    \setlength{\tabcolsep}{1pt}

    \begin{subfigure}[t]{\linewidth}
        \centering
        \begin{tabular}{cccc}
            \includegraphics[width=0.24\linewidth]{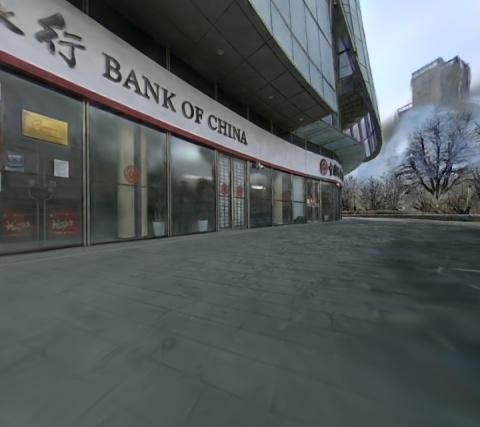} &
            \includegraphics[width=0.24\linewidth]{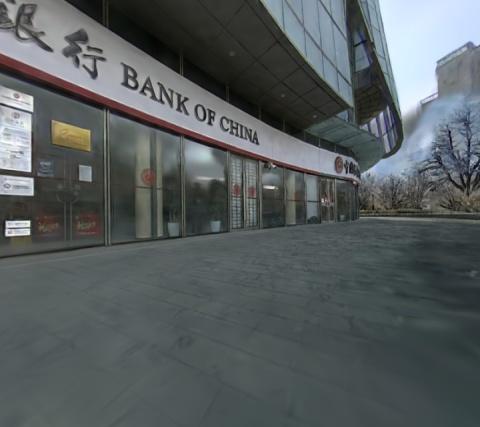} &
            \includegraphics[width=0.24\linewidth]{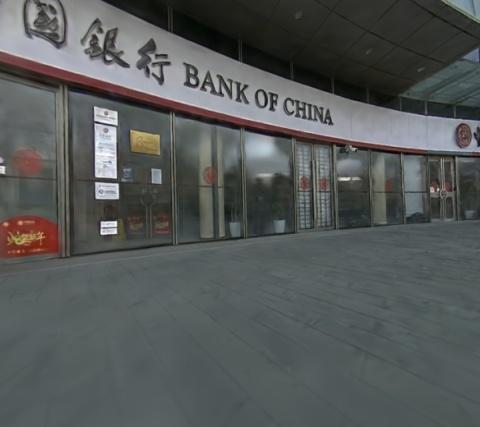} &
            \includegraphics[width=0.24\linewidth]{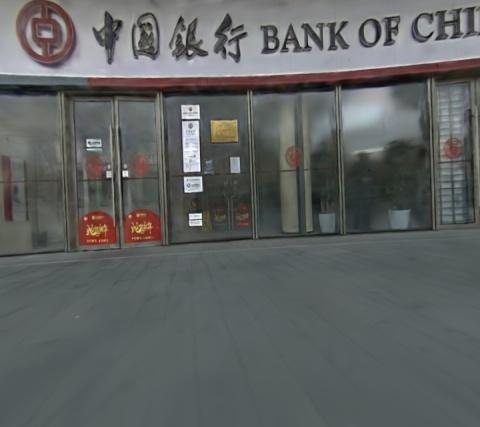} \\
            \includegraphics[width=0.24\linewidth]{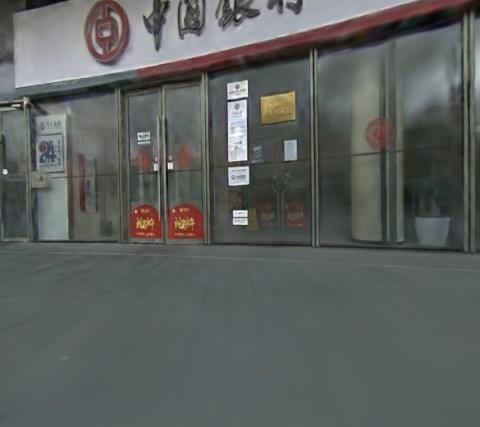} &
            \includegraphics[width=0.24\linewidth]{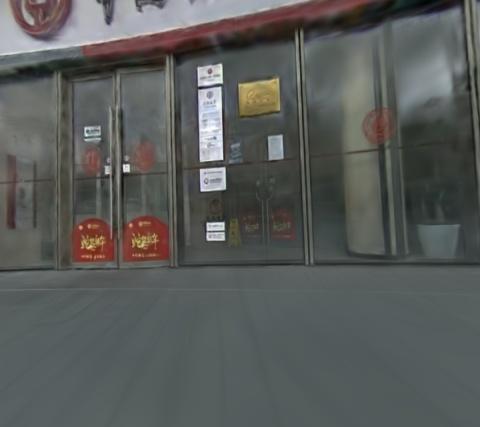} &
            \includegraphics[width=0.24\linewidth]{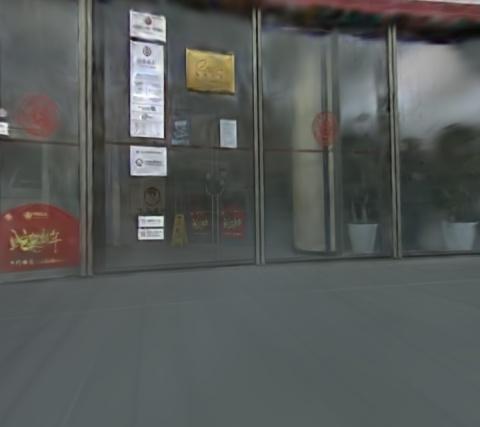} &
            \includegraphics[width=0.24\linewidth]{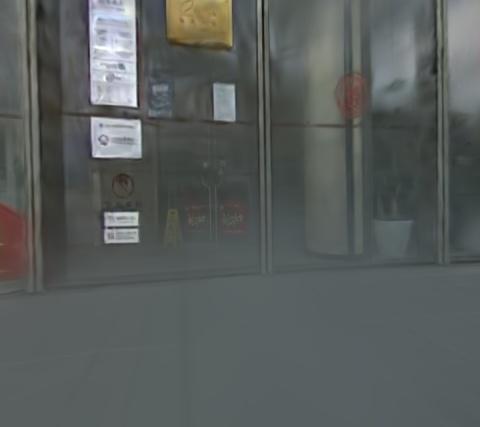}
        \end{tabular}
        \caption{\textbf{User instruction:} Navigate to \textit{Bank of China}.}
        \label{fig:demo_case1}
    \end{subfigure}

    \vspace{0.6ex}

    \begin{subfigure}[t]{\linewidth}
        \centering
        \begin{tabular}{cccc}
            \includegraphics[width=0.24\linewidth]{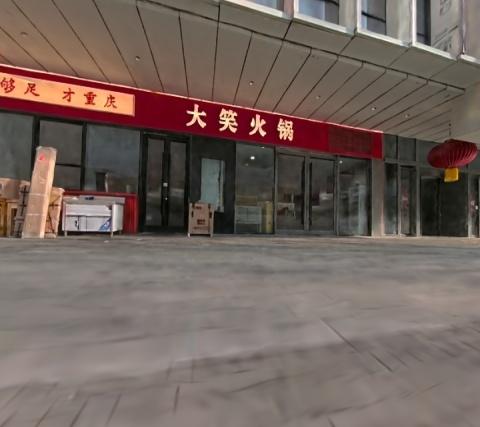} &
            \includegraphics[width=0.24\linewidth]{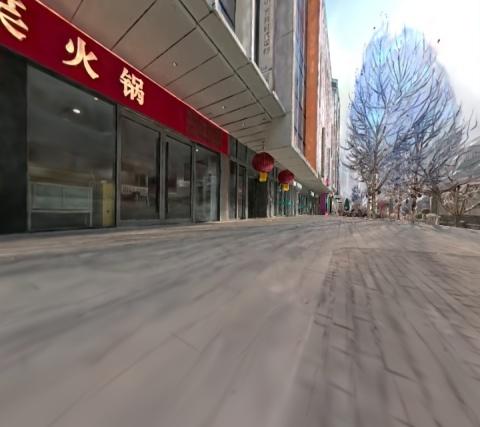} &
            \includegraphics[width=0.24\linewidth]{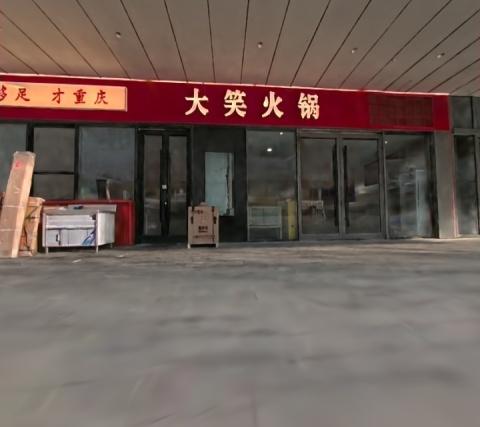} &
            \includegraphics[width=0.24\linewidth]{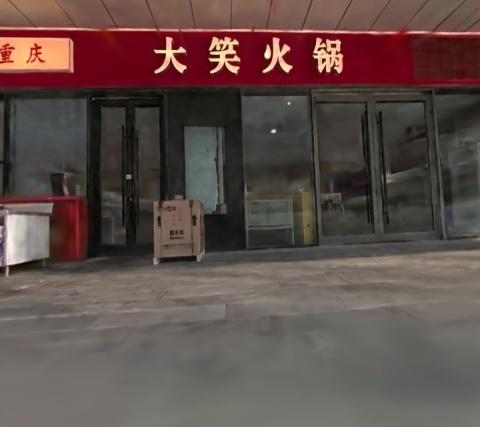} \\
            \includegraphics[width=0.24\linewidth]{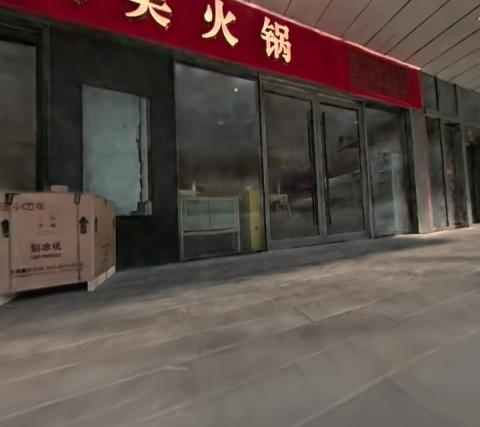} &
            \includegraphics[width=0.24\linewidth]{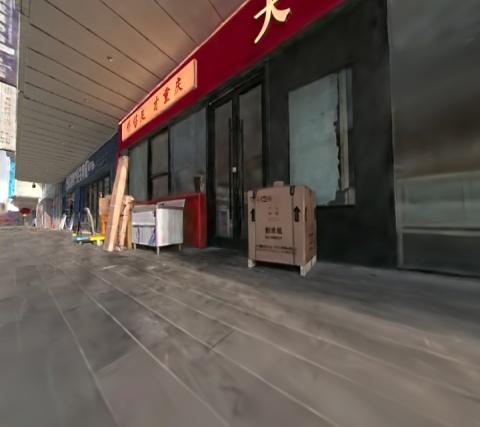} &
            \includegraphics[width=0.24\linewidth]{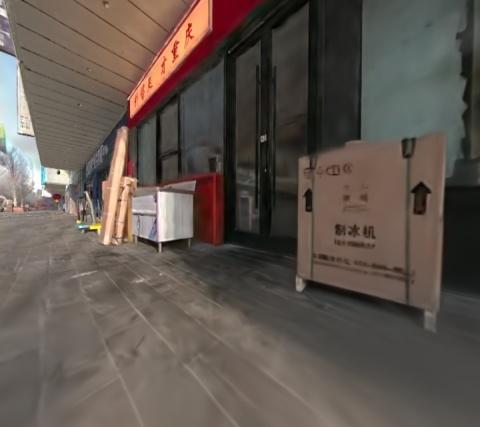} &
            \includegraphics[width=0.24\linewidth]{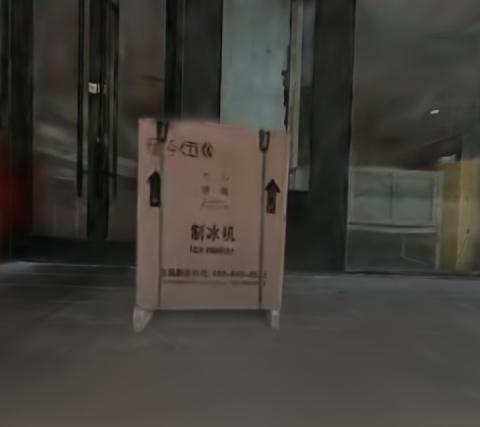}
        \end{tabular}
        \caption{\textbf{User instruction:} Go to \textit{Daxiao Hotpot}.}
        \label{fig:demo_case2}
    \end{subfigure}

    \caption{\textbf{Visualization of egocentric images during embodied navigation in POINav-Bench.}
    (a--b) show two navigation cases where the agent progressively approaches the target POI from a distant start to close proximity.}
    \label{fig:demo}
\end{figure}

In Figure \ref{fig:demo}, we illustrate how the egocentric view of the embodied agent changes during navigation with POINav on our proposed benchmark. From the figure, it can be observed that the agent is able to continuously approach the target in simulated outdoor environments until it reaches the entrance of the target POI. 

\begin{figure}[t!]
    \centering
    \includegraphics[width=\textwidth]{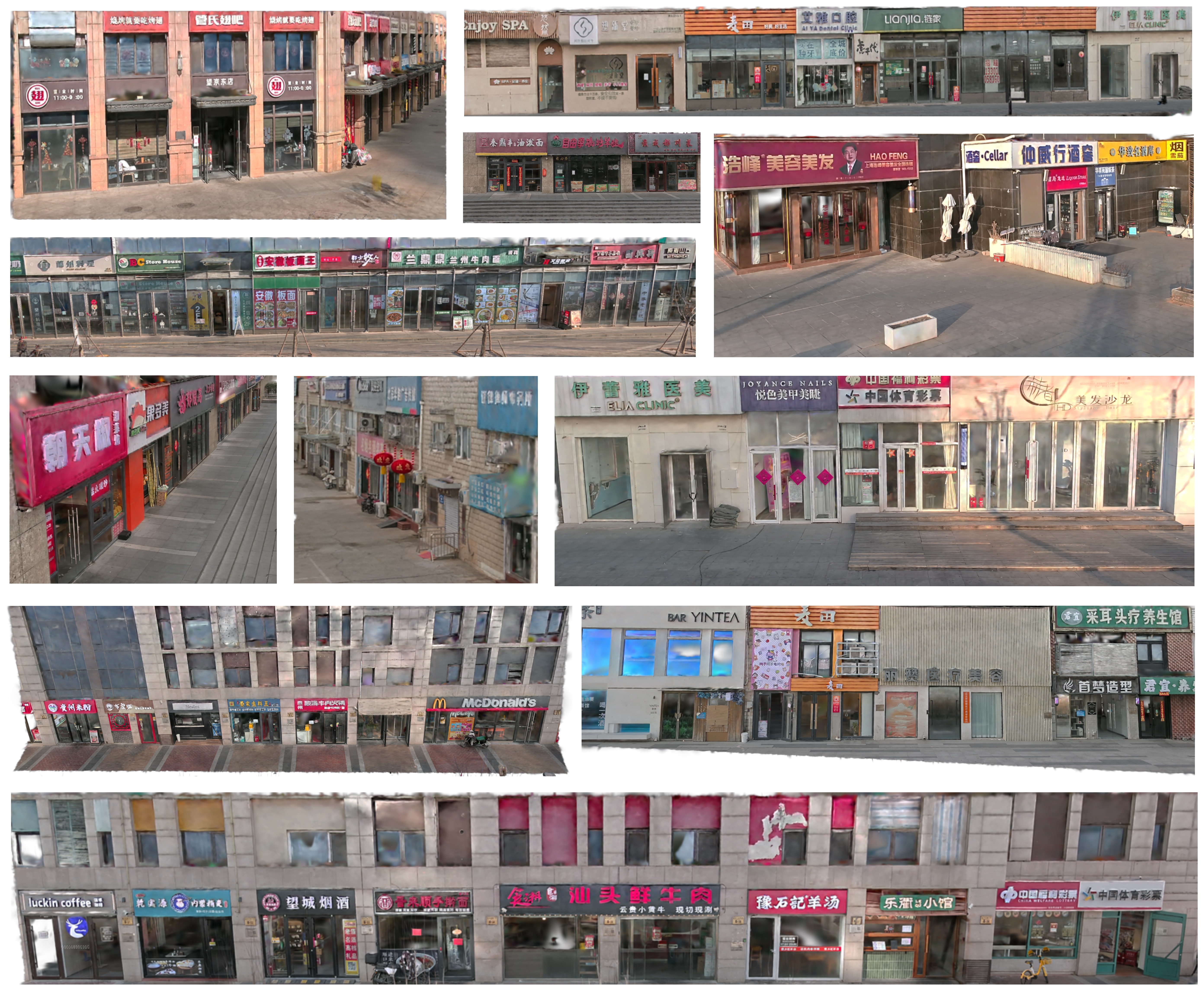}
    \caption{\textbf{Subsets of scenes in POINav-Bench.} Each 3DGS scene faithfully reproduces the storefront textures, legible signage, and spatial layouts of real-world commercial streets. The high density of visually similar, co-located POIs poses substantial challenges for precise entrance-level localization during navigation.}
    \label{fig:benchcase}
\end{figure}

\subsection{Failure Analysis}

We have presented two failure cases in Appendix~\ref{sec:supp-qualitative}. Our investigation reveals a common failure pattern across all BridgeNav-trained models: they fail completely when a large initial horizontal angular offset coincides with multiple visible POIs. In contrast, vanilla OmniNav (without BridgeNav training) can still succeed occasionally in such cases. For POINav, we find that while its brain model accurately grounds the target POI and provides visual context to the action model, the pipeline still breaks down at the downstream action control stage. Conversely, BridgeNav-trained models perform reliably on forward-view POIs with small deflections, where vanilla OmniNav underperforms.

We attribute this discrepancy to an inherent bias in the BridgeNav dataset, which is constructed from videos generated by Wan2.1-I2V. The dataset lacks sufficient trajectories with large horizontal angular offsets. Even in these challenging scenarios, the brain model still provides valid contextual guidance; however, this training distribution bias prevents POINav from executing effective action control toward the image goal.

\begin{figure}[t!]
    \centering
    \includegraphics[width=0.5\textwidth, height=6cm]{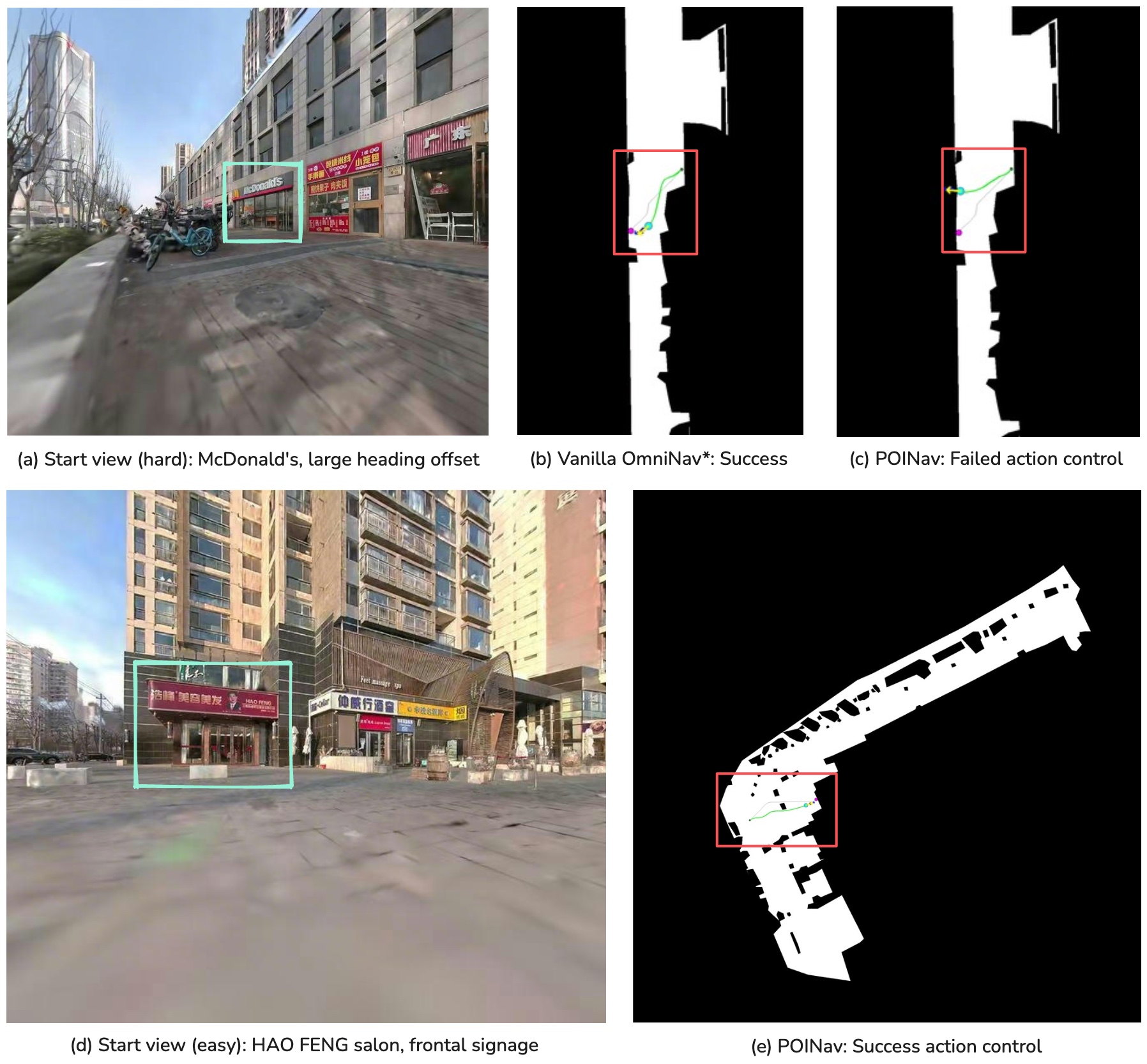}
    \caption{\textbf{POINav's failure and success patterns.} POINav fails in large-heading-offset scenarios but succeeds on forward-view POIs, whereas vanilla OmniNav exhibits the opposite trend.}
    \label{fig:failure_success_case}
\end{figure}

\subsection{POI-Grounded Reasoning Analysis}

\begin{table}[htbp]
\centering
\renewcommand{\arraystretch}{1.2} 
\begin{tabularx}{0.97\linewidth}{@{} l >{\centering\arraybackslash}X >{\centering\arraybackslash}X @{}} 
\toprule
\textbf{Metric / Category} & \textbf{Backbone} & \textbf{POINav (Ours)} \\
\midrule
\multicolumn{3}{l}{\textit{POI-Grounded Reasoning Metrics ($\uparrow$)}} \\
Referential Correctness (RC) & 91.6\% & \textbf{98.8\%} \\
Grounding Quality (GQ)       & 88.0\% & \textbf{94.8\%} \\
\midrule
\multicolumn{3}{l}{\textit{Failure Analysis ($\downarrow$)}} \\
Referential Error (RC \& GQ) & 8.4\% & \textbf{1.2\%} \\
Ambiguous Predictions (GQ)   & \textbf{3.6\%} & 4.0\% \\
\bottomrule
\end{tabularx}
\caption{Performance of the backbone VLM model and our POINav.}
\label{tab:poilocalization}
\end{table}

Since POINav follows the brain-action paradigm, a VLM-based “brain” module performs POI-grounded reasoning to localize the target entrance. This step critically determines the system’s upper bound.

We evaluate (i) \textbf{referential correctness (RC)}: whether the predicted bounding box refers to the queried POI, and (ii) \textbf{grounding quality (GQ)}: whether the predicted box ground one of the annotated valid entrances of the queried POI. Essentially, RC further indicates whether the visual input fed into the action model includes adequate referential visual cues. Meanwhile, GQ assesses whether the predicted bounding box is visually unambiguous, i.e., it does not contain any entrance of other POIs.

In our test set, each POI is annotated with a single entrance; however, in practice, many POIs such as shopping malls, hospitals, or multi-tenant buildings, feature multiple valid entrances that are visually distinct yet semantically equivalent. This inherent multiplicity renders conventional IoU-based metrics inadequate, as a prediction may correctly localize a valid but unannotated entrance and still be unfairly penalized. To address this challenge, we adopt a hybrid evaluation protocol that combines LLM-as-a-judge with human verification. Specifically, we visualize both the predicted bounding box and the human-annotated ground-truth box on the input image and feed this augmented visual context to a commercial VLM. The VLM is prompted to assess two binary judgments: (1) Referential Correctness (RC) is positive if the prediction points to any valid entrance of the queried POI; (2) Grounding Quality (GQ) is positive if the predicted bounding box reasonably covers that entrance without encompassing entrances of other POIs.

Models are evaluated on hundreds of samples, and Table \ref{tab:poilocalization} illustrates the evaluation result. The brain module of our POINav achieves 98.8\% referential correctness and 94.8\% grounding quality. Additionally, 4.0\% of the predictions are ambiguous: these predictions cover the target entrance of referential POI while also including at least one entrance of a nearby POI, making the visual evidence non-discriminative despite partial overlap with the correct target. 

The backbone model achieves only 91.6\% Referential Correctness (7.2\% lower than POINav). In terms of the action module, where each action relies on a specific visual grounding prediction, such per-step errors can accumulate rapidly, leading to drastically divergent navigation outcomes. This highlights the effectiveness of our POI-grounded reasoning in delivering stable visual cues that prevent error propagation to the action module.

The extremely high referential correctness indicates that our brain module can provide downstream action modules with reliable POI-specific visual cues in nearly all cases. From another perspective, only 1.2\% of predictions exhibit referential errors, and an additional 4.0\% produce visually ambiguous cues because a single predicted box covers entrances of multiple nearby POIs. Together, these are the main sources of POI-level visual ambiguity introduced by the brain module. Consequently, the brain module supplies high-quality and largely unambiguous visual guidance in approximately 95\% of samples, which suggests a very high performance ceiling for POINav on POI-goal navigation when coupled with a strong action module.



\section{Conclusion}
We present POINav-Bench, a POI-goal benchmark built from 11 real-world commercial areas reconstructed as 3D Gaussian Splatting (3DGS) scenes and integrated into Isaac Sim for closed-loop evaluation. We also propose the POINav Brain-Action Framework, which tackles final-meters POI navigation as by combining POI-grounded reasoning with Global-Context Action Querying. Experiments on POINav-Bench support the effectiveness of this Brain-Action formulation in 3DGS-based simulation environments. We hope this work can serve as a stepping stone toward more robust and precise POI-goal navigation in real-world environments.


\section{Limitations}
POINav-Bench currently covers 11 commercial areas, which limits geographic, cultural, and architectural diversity and may bias conclusions toward POI-dense retail settings. Our evaluation is conducted in 3DGS-based simulation, so sim-to-real gaps are not fully tackled. The benchmark and framework emphasize entrance-oriented POIs with signage-to-entrance grounding, which may not generalize to POIs without clear signage, ambiguous entrances, or temporally varying access conditions. Finally, reconstruction and annotation quality can affect both grounding and downstream action, and these failure modes warrant deeper analysis as end-to-end results mature.

\newpage

\appendix
\section*{Appendix}

\section{Benchmark Statistics and Setup}
\label{sec:supp-benchmark-stats}

\subsection{Data Statistics}

\begin{figure}[htbp]
\centering
\begin{minipage}[t]{0.6\linewidth}
\vspace{0pt}
\centering
\scriptsize
\setlength{\tabcolsep}{4pt}
\renewcommand{\arraystretch}{1.15}
\resizebox{\linewidth}{!}{%
\begin{tabular}{c c c c c c}
\toprule
\textbf{Scene} & \textbf{Area ($m^2$)} & \textbf{POIs} & \textbf{Gaussian Points} & \textbf{Mesh Vertices} & \textbf{Mesh Faces} \\
\midrule
Scene 1 & 7,825.44 & 10 & 3,325,648 & 247,282 & 384,545 \\
Scene 2 & 23,550.80 & 38 & 6,576,185 & 332,937 & 521,523 \\
Scene 3 & 15,326.98 & 35 & 4,728,910 & 403,478 & 618,515 \\
Scene 4 & 18,889.08 & 14 & 5,681,083 & 455,433 & 740,692 \\
Scene 5 & 17,782.10 & 14 & 4,080,656 & 317,538 & 515,075 \\
Scene 6 & 4,157.67 & 8 & 2,478,824 & 229,736 & 345,700 \\
Scene 7 & 7,584.83 & 10 & 2,459,165 & 149,118 & 225,390 \\
Scene 8 & 11,300.19 & 18 & 3,787,917 & 291,624 & 449,158 \\
Scene 9 & 7,469.99 & 9 & 1,868,671 & 123,310 & 196,551 \\
Scene 10 & 6,598.80 & 9 & 1,981,285 & 147,440 & 229,364 \\
Scene 11 & 5,912.50 & 6 & 1,871,173 & 112,688 & 175,698 \\
\midrule
Total & 126,398.38 & 163 & 38,839,517 & 2,810,584 & 4,402,211 \\
\bottomrule
\end{tabular}%
}
\captionsetup{font=footnotesize,justification=raggedright,singlelinecheck=false}
\captionof{table}{Scene-level statistics of POINav-Bench.}
\label{tab:scene_stats}
\end{minipage}
\hfill
\begin{minipage}[t]{0.355\linewidth}
\vspace{0pt}
\centering
\includegraphics[width=\linewidth]{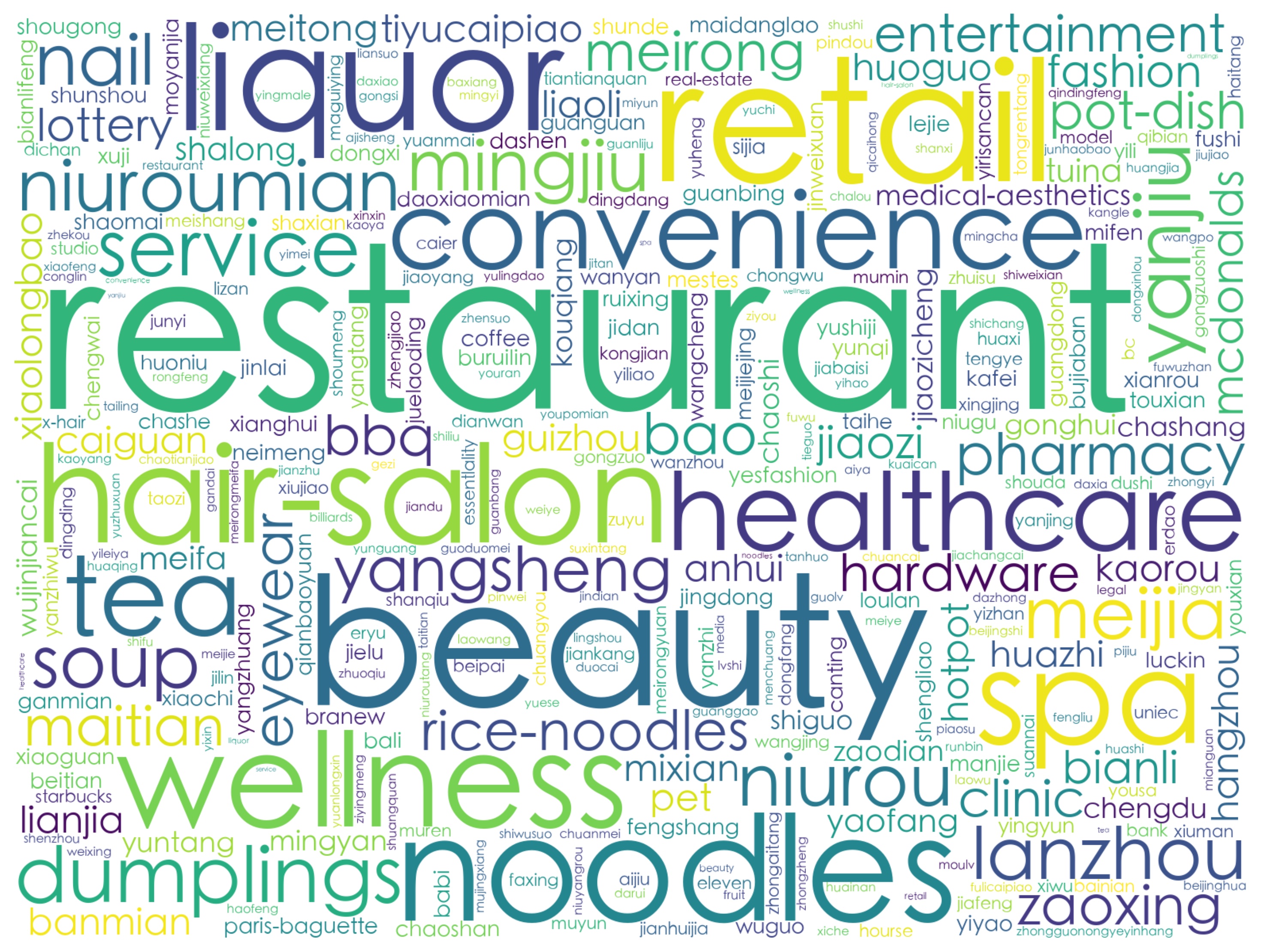}

\captionsetup{font=footnotesize,justification=raggedright,singlelinecheck=false}
\captionof{figure}{POI word cloud.}
\label{fig:poi_wordcloud}
\end{minipage}
\end{figure}

We present the detailed statistics of our dataset in Table~\ref{tab:scene_stats}. Our benchmark comprises 11 diverse scenes covering a total area of approximately 126,398 $m^2$. These scenes feature high-fidelity 3D Gaussian Splatting reconstructions with over 38 million Gaussian points and dense mesh geometry. Across these scenes, we have annotated 163 unique Points of Interest (POIs).

The distribution of POIs covers a wide range of functional categories, ensuring a comprehensive evaluation of navigation capabilities. Specifically, the dataset categorizes POIs into five primary classes: Dining, Retail, Medical \& Health, Service, and Entertainment. This taxonomy reflects real-world commercial and service-oriented environments, providing a challenging and realistic testbed for embodied agents. The word cloud in Figure~\ref{fig:poi_wordcloud} further highlights the diversity of specific POI names encountered in the scenes.

\subsection{Benchmark Comparison}

To contextualize the advances of POINav-Bench, we provide a systematic comparison with representative visual navigation benchmarks in Table~\ref{tab:benchmark_comparison}. Early work such as VLN-CE~\cite{krantz2020beyond} established a foundation for continuous embodied navigation, but it relies on scanned mesh environments that often contain noticeable geometric holes and relatively limited rendering fidelity. More recent efforts such as SAGE-Bench~\cite{miao2025physicallyexecutable3dgaussian} improve visual realism through 3D Gaussian Splatting, yet they primarily target object-level navigation rather than the more demanding entrance-level POI arrival setting considered in our benchmark.

The distinction becomes clearer in urban-scene benchmarks. CitySeeker~\cite{wang2025cityseeker} uses real-world panoramas, but its graph-based evaluation protocol constrains the agent to transitions between sparse nodes, which limits assessment of precise final-meter behavior. In contrast, POINav-Bench provides a continuous navigation space in which the agent must execute fine-grained motion control to reach the target entrance accurately.

This gap is even more pronounced in recent POI-oriented benchmarks such as BridgeNav~\cite{zhao2026bridgingindooroutdoorgapvisioncentric} and ABot-N0~\cite{chu2026abot}, which adopt open-loop evaluation on generated videos. Such settings often contain visual distortions and lack the interactive feedback required for realistic policy evaluation. POINav-Bench addresses these limitations by combining real-world captures with a closed-loop simulator built on Isaac Sim, enabling physically grounded interaction and collision-aware evaluation. Moreover, instead of using 2D pixel-level targets, our benchmark provides high-quality 3D bounding box annotations for entrances, allowing agents to be evaluated on physically precise arrival at functional access points in a photorealistic environment.

\begin{table}[t!]
\centering
\footnotesize
\setlength{\tabcolsep}{4pt}
\renewcommand{\arraystretch}{1.12}
\resizebox{\linewidth}{!}{%
\begin{tabular}{@{} c c c c c c c @{} }
\toprule
\textbf{Benchmark} & \textbf{Loop Type} & \textbf{Nav. Space} & \shortstack{\textbf{Target}\\\textbf{Granularity}} & \shortstack{\textbf{Entrance}\\\textbf{Label}} & \textbf{Physics} & \shortstack{\textbf{Visual}\\\textbf{Fidelity}} \\
\midrule
VLN-CE~\cite{krantz2020beyond} & Closed-loop & Continuous & Point / Object & None & Basic & Low \\
SAGE-Bench~\cite{miao2025physicallyexecutable3dgaussian} & Closed-loop & Continuous & Object-level & None & Basic & High \\
CitySeeker~\cite{wang2025cityseeker} & Closed-loop & Graph-based & Street-level & None & None & Photo-real \\
BridgeNav~\cite{zhao2026bridgingindooroutdoorgapvisioncentric} & Open-loop & Video-based & Entrance-level & 2D Pixel & None & Distorted \\
ABot-N0~\cite{chu2026abot} & Open-loop & Video-based & Entrance-level & 2D Pixel & None & Distorted \\
POINav (Ours) & Closed-loop & Continuous & Entrance-level & 3D BBox & Isaac Sim & Photorealistic \\
\bottomrule
\end{tabular}%
}
\caption{Comparison with representative VLN benchmarks and POI-goal evaluation settings along the axes most relevant to final-meters arrival.}
\label{tab:benchmark_comparison}
\end{table}

\subsection{Reconstruction Setup}
\label{sec:supp-recon-sim}

Each of the 11 commercial areas in POINav-Bench is reconstructed from high-precision LiDAR point clouds and photogrammetry data collected on-site. To ensure both visual fidelity and physical accuracy, we adopt a specialized four-phase pipeline that fuses geometric constraints with radiometric information, as outlined in Algorithm~1.

\begin{wrapfigure}{R}{0.52\linewidth}
\vspace{-12pt}
\centering
\small
\renewcommand{\arraystretch}{1.1}
\resizebox{\linewidth}{!}{%
\begin{tabular}{@{}l@{}}
\toprule
\textbf{Algorithm 1:} Reconstruction Pipeline \\
\midrule
\textbf{Input:} LiDAR $\mathcal{P}$, RGB $\mathcal{I}$, Aux.\ $\mathcal{A}$, IMU $\mathcal{U}$ \\
\textbf{Output:} 3DGS Scene $\mathcal{S}$, Collision Mesh $\mathcal{M}$ \\
\addlinespace[0.06cm]
\textcolor{gray}{\textit{// Phase 1: LiDAR-Inertial Mapping}} \\
1:\; $T_{init},\mathcal{P}_{raw} \!\leftarrow\! \textbf{LI-SLAM}(\mathcal{P},\mathcal{U},\mathcal{A})$ \\
\addlinespace[0.06cm]
\textcolor{gray}{\textit{// Phase 2: Global Pose Refinement}} \\
2:\; $T_{ref},\mathcal{P}_{aln} \!\leftarrow\! \textbf{GlobalOpt}(T_{init},\mathcal{P}_{raw},\mathcal{A})$ \\
\addlinespace[0.06cm]
\textcolor{gray}{\textit{// Phase 3: LiDAR-Init 3DGS Training}} \\
3:\; $G_{init} \!\leftarrow\! \textbf{InitGS}(\mathcal{P}_{aln})$ \\
4:\; $\mathcal{S} \!\leftarrow\! \textbf{Train3DGS}(G_{init},T_{ref},\mathcal{I})$ \\
\addlinespace[0.06cm]
\textcolor{gray}{\textit{// Phase 4: Physics Asset Extraction}} \\
5:\; $\mathcal{M} \!\leftarrow\! \textbf{ExtractMesh}(\mathcal{S})$ \\
\bottomrule
\end{tabular}%
}
\vspace{-6pt}
\end{wrapfigure}

\begin{figure}[t]
	\centering
	\includegraphics[width=\linewidth]{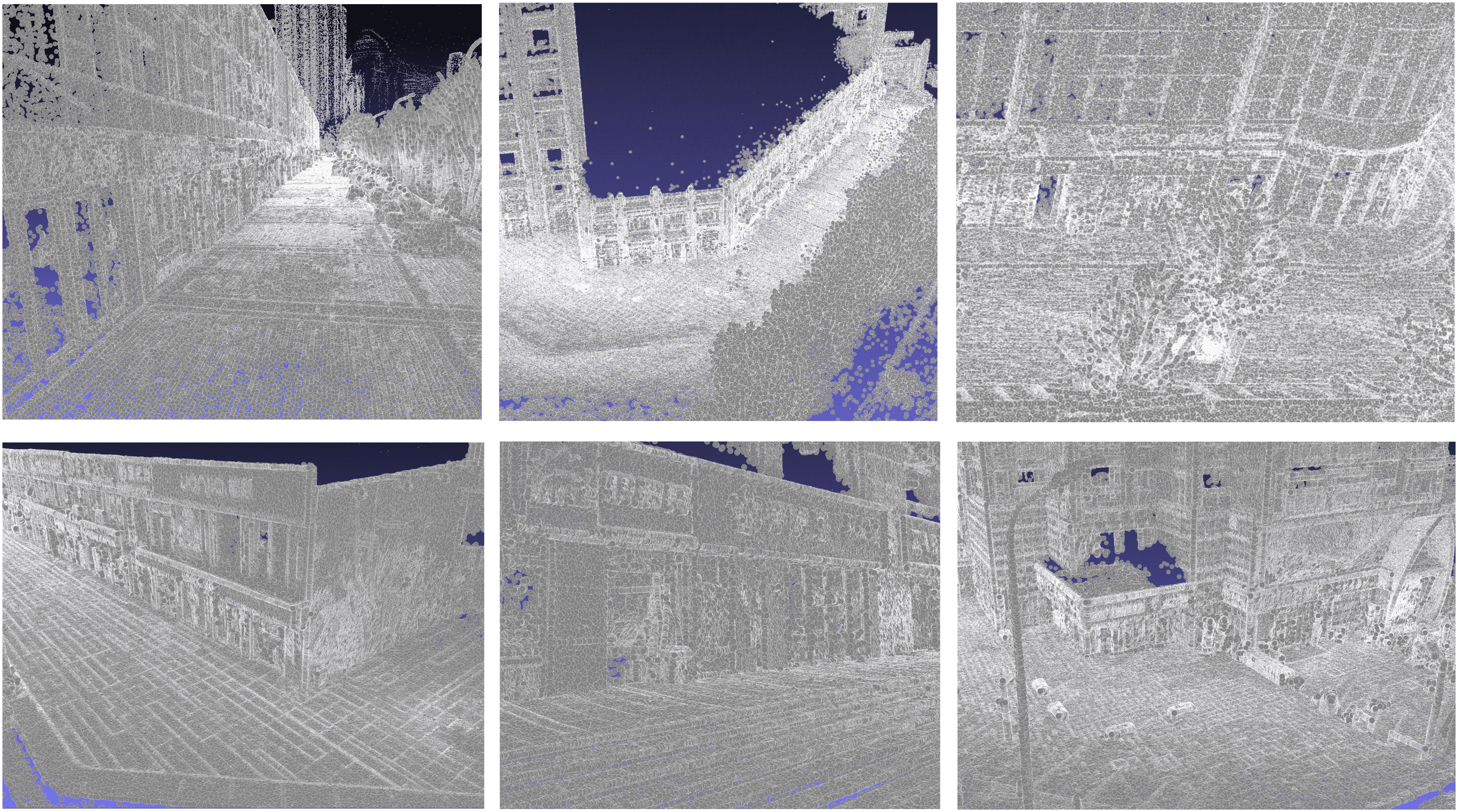}
	\caption{Extracted mesh geometry from 3DGS reconstructions of representative scenes in POINav-Bench. The meshes exhibit high geometric fidelity, capturing fine architectural details essential for physically grounded navigation evaluation.}
	\label{fig:3dmesh}
\end{figure}

In \textbf{Phase~1}, a multi-sensor capture platform performs real-time LiDAR-Inertial SLAM during on-site data collection, producing a coarse trajectory and an initial point cloud map. \textbf{Phase~2} then transfers the raw data to a compute server for global multi-sensor joint optimization, yielding a refined trajectory and a dense, well-aligned point cloud registered with the RGB image frames. Building on the refined geometry, \textbf{Phase~3} initializes 3D Gaussian Splatting directly from the aligned dense LiDAR point cloud rather than sparse SfM points. Combined with the refined camera poses, this strategy significantly improves the geometric accuracy of the resulting scene representation. Finally, \textbf{Phase~4} extracts a dense triangle mesh from the trained Gaussians to serve as the collision geometry for physically grounded evaluation in Isaac Sim, as shown in Figure~\ref{fig:3dmesh}.

\section{Additional Qualitative Analysis}
\label{sec:supp-qualitative}

A central contribution of POINav-Bench is its photorealistic visual fidelity, which bridges the sim-to-real gap that undermines evaluation on synthetic or generated scenes. Crucially, all episodes are evaluated in a closed-loop setting within Isaac Sim: the agent must react to its own previous actions and accumulating observation history in real time, rather than being scored on pre-recorded trajectories. This interactive protocol amplifies the impact of visual fidelity, since subtle rendering details directly influence the agent's sequential decision-making.

\begin{figure}[t!]
\centering
\includegraphics[width=0.245\linewidth]{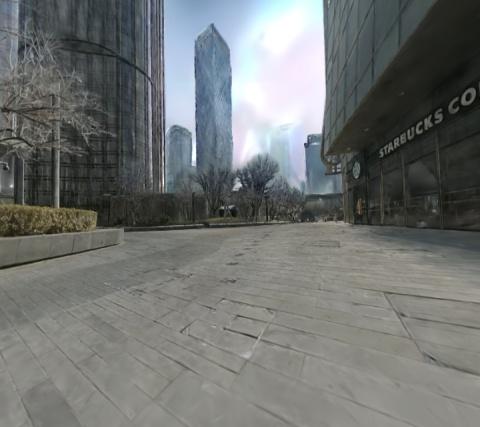}\hfill
\includegraphics[width=0.245\linewidth]{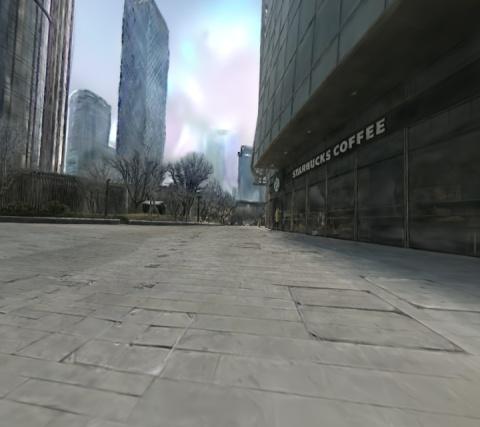}\hfill
\includegraphics[width=0.245\linewidth]{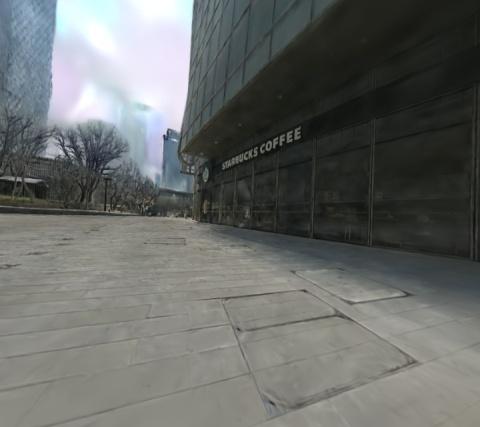}\hfill
\includegraphics[width=0.245\linewidth]{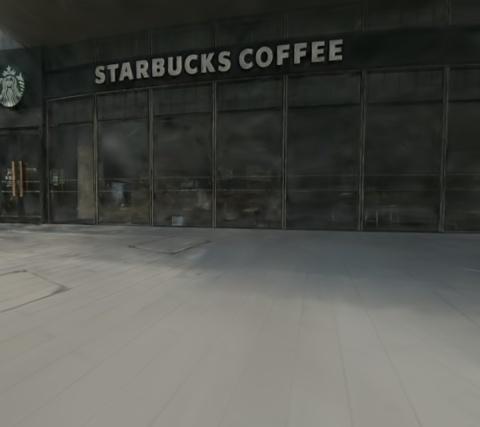}\\[2pt]
{\centering\small (a) Case 1: Near-success at Starbucks Coffee\par}
\vspace{6pt}
\includegraphics[width=0.245\linewidth]{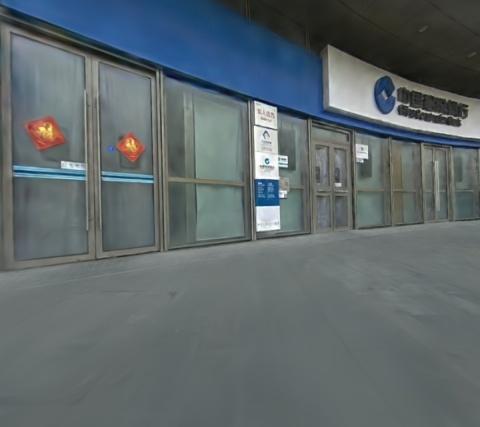}\hfill
\includegraphics[width=0.245\linewidth]{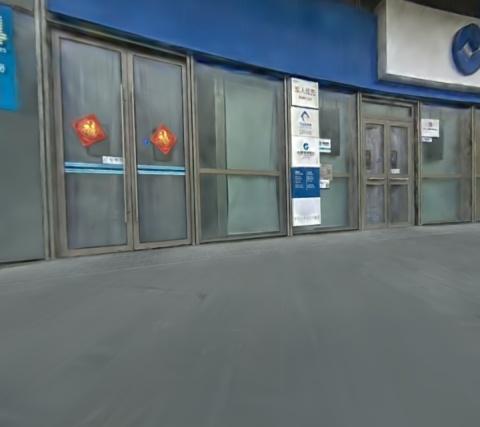}\hfill
\includegraphics[width=0.245\linewidth]{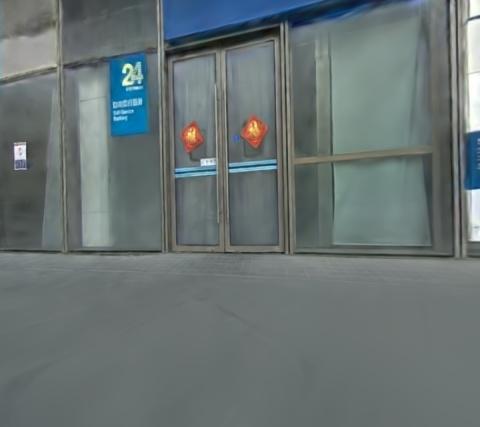}\hfill
\includegraphics[width=0.245\linewidth]{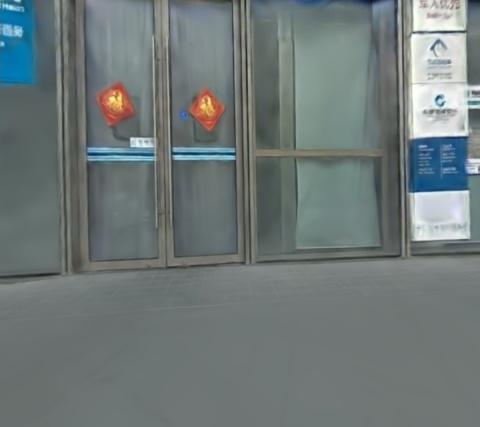}\\[2pt]
{\centering\small (b) Case 2: Multi-entrance ambiguity at China Construction Bank\par}
\caption{Representative navigation episodes illustrating photorealistic benchmark fidelity. Each row shows four sequential egocentric frames (left to right). \textbf{(a)}~The agent correctly approaches the Starbucks storefront but terminates slightly offset from the entrance due to glass-facade visual ambiguity. \textbf{(b)}~The agent reaches the correct bank building but converges on the 24h self-service entrance rather than the annotated personal banking entrance due to near-identical frosted-glass door appearances.}
\label{fig:qualitative_cases}
\end{figure}

It is worth noting that the action module of our POINav framework is trained entirely on the BridgeNav~\cite{zhao2026bridgingindooroutdoorgapvisioncentric} navigation dataset, which consists of video sequences synthesized by a generative model. The action policy has therefore never been exposed to real-world or 3DGS-rendered imagery during training. Despite this domain gap, the following two cases (Figure~\ref{fig:qualitative_cases}) show that the agent navigates reasonably in our photorealistic scenes, reaching the close vicinity of the target POIs. The remaining errors, however, reveal visual challenges that are well captured by our high-fidelity reconstruction and would be absent in lower-quality benchmarks.

\paragraph{Case 1: Near-Success at Starbucks Coffee.}
Figure~\ref{fig:qualitative_cases}(a) depicts a representative navigation episode where the agent is instructed to navigate to \textit{Starbucks Coffee}. The four egocentric frames, captured from the 3DGS-rendered scene in Isaac Sim, demonstrate the progressive approach trajectory. In the initial frames, the agent correctly identifies the Starbucks signage at a distance and begins heading toward the storefront along a wide commercial sidewalk. The rendered scene accurately reproduces urban details such as reflective glass curtain walls, bare winter trees, pavement tile patterns, and the characteristic green logo and ``STARBUCKS COFFEE'' lettering on the facade.

As the agent approaches, the visual complexity intensifies: the low camera mounting height of the quadrupedal robot causes the high-mounted signage to gradually exit the vertical field of view, while the glass storefront increasingly dominates the observation. In a closed-loop setting, each waypoint prediction is conditioned on the agent's own evolving viewpoint, and this progressive loss of the semantic anchor creates compounding uncertainty in the final meters. In the last frame, the agent arrives in the immediate vicinity of the entrance but terminates slightly offset from the annotated entrance bounding box.

Notably, the action policy driving this episode has only been trained on synthetically generated video sequences, yet it still manages to approach the correct storefront and reach near-success. The residual offset can be attributed to the photorealistic visual ambiguity of the glass-facade storefront, where the entrance boundary is nearly indistinguishable from adjacent non-entrance glass panels---a subtle geometric challenge preserved by our 3DGS reconstruction that would be absent in lower-fidelity benchmarks.

\paragraph{Case 2: Multi-Entrance Ambiguity at China Construction Bank.}
Figure~\ref{fig:qualitative_cases}(b) presents a failure episode targeting \textit{China Construction Bank}. The 3DGS-rendered scene reproduces the detailed facade of a large bank branch: a distinctive blue awning, frosted glass double-doors decorated with red Spring Festival couplets, small blue directional signs (e.g., ``24h Self-Service Banking'', ``Personal Banking''), and the bank's corporate logo mounted on the upper-right corner. These visual cues are well preserved by the photorealistic reconstruction.

The navigation challenge in this case arises from multi-entrance ambiguity. As is common for real-world bank branches, the storefront features several visually similar frosted-glass door panels spanning a wide facade, with only subtle signage differences distinguishing the main personal banking entrance from auxiliary access points such as the 24-hour self-service area. In the first two frames, the agent approaches a door decorated with red couplets and a blue ``24'' sign overhead, which corresponds to the self-service banking entrance rather than the annotated main entrance. As the agent continues laterally along the facade (frames 3--4), the corporate logo and ``Personal Banking'' sign become visible on the right side, but the agent has already committed to the wrong door segment.

In a closed-loop evaluation, this type of early commitment is particularly consequential: once the agent adjusts its heading toward the self-service door, subsequent observations reinforce this trajectory, making mid-course correction increasingly unlikely without explicit re-grounding. Again, given that the action policy is trained exclusively on synthetic data and has never encountered real-world bank facades, the fact that it successfully navigates to the correct building and localizes a plausible entrance is encouraging. The remaining error stems from entrance disambiguation---distinguishing functionally different access points that share near-identical visual appearances (frosted glass, identical door frames, similar signage styles). This type of challenge is characteristic of real-world POI-goal navigation, one that our photorealistic benchmark preserves and that lower-fidelity or generated-scene evaluations would obscure.

\clearpage

%
%
\bibliographystyle{splncs04}
\bibliography{main}

\begin{thebibliography}{10}
\providecommand{\url}[1]{\texttt{#1}}
\providecommand{\urlprefix}{URL }
\providecommand{\doi}[1]{https://doi.org/#1}

\bibitem{anderson2018vision}
Anderson, P., Wu, Q., Teney, D., Bruce, J., Johnson, M., S{\"u}nderhauf, N., Reid, I., Gould, S., Van Den~Hengel, A.: Vision-and-language navigation: Interpreting visually-grounded navigation instructions in real environments. In: Proceedings of the IEEE conference on computer vision and pattern recognition. pp. 3674--3683 (2018)

\bibitem{bai2025qwen3vltechnicalreport}
Bai, S., Cai, Y., Chen, R., Chen, K., Chen, X., Cheng, Z., Deng, L., Ding, W., Gao, C., Ge, C., Ge, W., Guo, Z., Huang, Q., Huang, J., Huang, F., Hui, B., Jiang, S., Li, Z., Li, M., Li, M., Li, K., Lin, Z., Lin, J., Liu, X., Liu, J., Liu, C., Liu, Y., Liu, D., Liu, S., Lu, D., Luo, R., Lv, C., Men, R., Meng, L., Ren, X., Ren, X., Song, S., Sun, Y., Tang, J., Tu, J., Wan, J., Wang, P., Wang, P., Wang, Q., Wang, Y., Xie, T., Xu, Y., Xu, H., Xu, J., Yang, Z., Yang, M., Yang, J., Yang, A., Yu, B., Zhang, F., Zhang, H., Zhang, X., Zheng, B., Zhong, H., Zhou, J., Zhou, F., Zhou, J., Zhu, Y., Zhu, K.: Qwen3-vl technical report (2025), \url{https://arxiv.org/abs/2511.21631}

\bibitem{bai2025qwen25vltechnicalreport}
Bai, S., Chen, K., Liu, X., Wang, J., Ge, W., Song, S., Dang, K., Wang, P., Wang, S., Tang, J., Zhong, H., Zhu, Y., Yang, M., Li, Z., Wan, J., Wang, P., Ding, W., Fu, Z., Xu, Y., Ye, J., Zhang, X., Xie, T., Cheng, Z., Zhang, H., Yang, Z., Xu, H., Lin, J.: Qwen2.5-vl technical report (2025), \url{https://arxiv.org/abs/2502.13923}

\bibitem{batra2020objectnav}
Batra, D., Gokaslan, A., Kembhavi, A., Maksymets, O., Mottaghi, R., Savva, M., Toshev, A., Wijmans, E.: Objectnav revisited: On evaluation of embodied agents navigating to objects. arXiv preprint arXiv:2006.13171  (2020)

\bibitem{bono2023end}
Bono, G., Antsfeld, L., Chidlovskii, B., Weinzaepfel, P., Wolf, C.: End-to-end (instance)-image goal navigation through correspondence as an emergent phenomenon. arXiv preprint arXiv:2309.16634  (2023)

\bibitem{chaplot2020object}
Chaplot, D.S., Gandhi, D.P., Gupta, A., Salakhutdinov, R.R.: Object goal navigation using goal-oriented semantic exploration. Advances in Neural Information Processing Systems  \textbf{33},  4247--4258 (2020)

\bibitem{chen2025socialnav}
Chen, Z., Guo, Y., Chu, Z., Luo, M., Shen, Y., Sun, M., Hu, J., Xie, S., Yang, K., Shi, P., et~al.: Socialnav: Training human-inspired foundation model for socially-aware embodied navigation. arXiv preprint arXiv:2511.21135  (2025)

\bibitem{cheng2024navila}
Cheng, A.C., Ji, Y., Yang, Z., Gongye, Z., Zou, X., Kautz, J., B{\i}y{\i}k, E., Yin, H., Liu, S., Wang, X.: Navila: Legged robot vision-language-action model for navigation. arXiv preprint arXiv:2412.04453  (2024)

\bibitem{chu2026abot}
Chu, Z., Xie, S., Wu, X., Shen, Y., Luo, M., Wang, Z., Liu, F., Leng, X., Hu, J., Yin, M., et~al.: Abot-n0: Technical report on the vla foundation model for versatile embodied navigation. arXiv preprint arXiv:2602.11598  (2026)

\bibitem{dorbala2022clip}
Dorbala, V.S., Sigurdsson, G., Piramuthu, R., Thomason, J., Sukhatme, G.S.: Clip-nav: Using clip for zero-shot vision-and-language navigation. arXiv preprint arXiv:2211.16649  (2022)

\bibitem{guo2025seed15vltechnicalreport}
Guo, D., Wu, F., Zhu, F., Leng, F., Shi, G., Chen, H., Fan, H., Wang, J., Jiang, J., Wang, J., Chen, J., Huang, J., Lei, K., Yuan, L., Luo, L., Liu, P., Ye, Q., Qian, R., Yan, S., Zhao, S., Peng, S., Li, S., Yuan, S., Wu, S., Cheng, T., Liu, W., Wang, W., Zeng, X., Liu, X., Qin, X., Ding, X., Xiao, X., Zhang, X., Zhang, X., Xiong, X., Peng, Y., Chen, Y., Li, Y., Hu, Y., Lin, Y., Hu, Y., Zhang, Y., Wu, Y., Li, Y., Liu, Y., Ling, Y., Qin, Y., Wang, Z., He, Z., Zhang, A., Yi, B., Liao, B., Huang, C., Zhang, C., Deng, C., Deng, C., Lin, C., Yuan, C., Li, C., Gou, C., Lou, C., Wei, C., Liu, C., Li, C., Zhu, D., Zhong, D., Li, F., Zhang, F., Wu, G., Li, G., Xiao, G., Lin, H., Yang, H., Wang, H., Ji, H., Hao, H., Shen, H., Li, H., Li, J., Wu, J., Zhu, J., Jiao, J., Feng, J., Chen, J., Duan, J., Liu, J., Zeng, J., Tang, J., Sun, J., Chen, J., Long, J., Feng, J., Zhan, J., Fang, J., Lu, J., Hua, K., Liu, K., Shen, K., Zhang, K., Shen, K., Wang, K., Pan, K., Zhang, K., Li, K., Li, L., Li, L., Shi, L., Han, L., Xiang, L., Chen, L., Chen, L., Li, L., Yan, L., Chi, L., Liu, L., Du, M., Wang, M., Pan, N., Chen, P., Chen, P., Wu, P., Yuan, Q., Shuai, Q., Tao, Q., Zheng, R., Zhang, R., Zhang, R., Wang, R., Yang, R., Zhao, R., Xu, S., Liang, S., Yan, S., Zhong, S., Cao, S., Wu, S., Liu, S., Chang, S., Cai, S., Ao, T., Yang, T., Zhang, T., Zhong, W., Jia, W., Weng, W., Yu, W., Huang, W., Zhu, W., Yang, W., Wang, W., Long, X., Yin, X., Li, X., Zhu, X., Jia, X., Zhang, X., Liu, X., Zhang, X., Yang, X., Luo, X., Chen, X., Zhong, X., Xiao, X., Li, X., Wu, Y., Wen, Y., Du, Y., Zhang, Y., Ye, Y., Wu, Y., Liu, Y., Yue, Y., Zhou, Y., Yuan, Y., Xu, Y., Yang, Y., Zhang, Y., Fang, Y., Li, Y., Ren, Y., Xiong, Y., Hong, Z., Wang, Z., Sun, Z., Wang, Z., Cai, Z., Zha, Z., An, Z., Zhao, Z., Xu, Z., Chen, Z., Wu, Z., Zheng, Z., Wang, Z., Huang, Z., Zhu, Z., Song, Z.: Seed1.5-vl technical report (2025), \url{https://arxiv.org/abs/2505.07062}

\bibitem{huang2026ticvlathinkincontrolvisionlanguageactionmodel}
Huang, Z., Zhang, Y., Liu, J., Song, R., Tang, C., Ma, J.: Tic-vla: A think-in-control vision-language-action model for robot navigation in dynamic environments (2026), \url{https://arxiv.org/abs/2602.02459}

\bibitem{ieong2025multimodal}
Ieong, I.T., Tang, H.: Multimodal perception for goal-oriented navigation: A survey. arXiv preprint arXiv:2504.15643  (2025)

\bibitem{kerbl20233d}
Kerbl, B., Kopanas, G., Leimk{\"u}hler, T., Drettakis, G., et~al.: 3d gaussian splatting for real-time radiance field rendering. ACM Trans. Graph.  \textbf{42}(4),  139--1 (2023)

\bibitem{krantz2022instance}
Krantz, J., Lee, S., Malik, J., Batra, D., Chaplot, D.S.: Instance-specific image goal navigation: Training embodied agents to find object instances. arXiv preprint arXiv:2211.15876  (2022)

\bibitem{krantz2020beyond}
Krantz, J., Wijmans, E., Majumdar, A., Batra, D., Lee, S.: Beyond the nav-graph: Vision-and-language navigation in continuous environments. In: European Conference on Computer Vision. pp. 104--120. Springer (2020)

\bibitem{ku2020room}
Ku, A., Anderson, P., Patel, R., Ie, E., Baldridge, J.: Room-across-room: Multilingual vision-and-language navigation with dense spatiotemporal grounding. In: Proceedings of the 2020 Conference on Empirical Methods in Natural Language Processing (EMNLP). pp. 4392--4412 (2020)

\bibitem{lin2025vlnversebenchmarkvisionlanguagenavigation}
Lin, S., Li, Z., Zhao, X., Zhou, G., Wang, L., Wei, R., Tang, R., Li, J., Wang, H., Pang, J., van~den Hengel, A., Liu, J., Wu, Q.: Vlnverse: A benchmark for vision-language navigation with versatile, embodied, realistic simulation and evaluation (2025), \url{https://arxiv.org/abs/2512.19021}

\bibitem{liu2025citywalker}
Liu, X., Li, J., Jiang, Y., Sujay, N., Yang, Z., Zhang, J., Abanes, J., Zhang, J., Feng, C.: Citywalker: Learning embodied urban navigation from web-scale videos. In: Proceedings of the Computer Vision and Pattern Recognition Conference. pp. 6875--6885 (2025)

\bibitem{long2024instructnav}
Long, Y., Cai, W., Wang, H., Zhan, G., Dong, H.: Instructnav: Zero-shot system for generic instruction navigation in unexplored environment. arXiv preprint arXiv:2406.04882  (2024)

\bibitem{miao2025physicallyexecutable3dgaussian}
Miao, B., Wei, R., Ge, Z., sun, X., Gao, S., Zhu, J., Wang, R., Tang, S., Xiao, J., Tang, R., Li, J.: Towards physically executable 3d gaussian for embodied navigation (2025), \url{https://arxiv.org/abs/2510.21307}

\bibitem{NVIDIA_Isaac_Sim}
{NVIDIA}: {Isaac Sim}, \url{https://github.com/isaac-sim/IsaacSim}

\bibitem{shah2023vint}
Shah, D., Sridhar, A., Dashora, N., Stachowicz, K., Black, K., Hirose, N., Levine, S.: Vint: A foundation model for visual navigation. arXiv preprint arXiv:2306.14846  (2023)

\bibitem{vteam2026glm45vglm41vthinkingversatilemultimodal}
Team, V., Hong, W., Yu, W., Gu, X., Wang, G., Gan, G., Tang, H., Cheng, J., Qi, J., Ji, J., Pan, L., Duan, S., Wang, W., Wang, Y., Cheng, Y., He, Z., Su, Z., Yang, Z., Pan, Z., Zeng, A., Wang, B., Chen, B., Shi, B., Pang, C., Zhang, C., Yin, D., Yang, F., Chen, G., Li, H., Zhu, J., Chen, J., Xu, J., Xu, J., Chen, J., Lin, J., Chen, J., Wang, J., Chen, J., Lei, L., Gong, L., Pan, L., Liu, M., Xu, M., Zhang, M., Zheng, Q., Lyu, R., Tu, S., Yang, S., Meng, S., Zhong, S., Huang, S., Zhao, S., Xue, S., Zhang, T., Luo, T., Hao, T., Tong, T., Jia, W., Li, W., Liu, X., Zhang, X., Lyu, X., Zhang, X., Fan, X., Huang, X., Xue, Y., Wang, Y., Wang, Y., Wang, Y., An, Y., Du, Y., Huang, Y., Niu, Y., Shi, Y., Wang, Y., Wang, Y., Yue, Y., Li, Y., Liu, Y., Zhang, Y., Wang, Y., Zhang, Y., Xue, Z., Du, Z., Hou, Z., Wang, Z., Zhang, P., Liu, D., Xu, B., Li, J., Huang, M., Dong, Y., Tang, J.: Glm-4.5v and glm-4.1v-thinking: Towards versatile multimodal reasoning with scalable reinforcement learning (2026), \url{https://arxiv.org/abs/2507.01006}

\bibitem{wang2025moge}
Wang, R., Xu, S., Dai, C., Xiang, J., Deng, Y., Tong, X., Yang, J.: Moge: Unlocking accurate monocular geometry estimation for open-domain images with optimal training supervision. In: Proceedings of the IEEE/CVF Conference on Computer Vision and Pattern Recognition. pp. 5261--5271 (2025)

\bibitem{wang2025cityseeker}
Wang, S., Liang, C., Gao, Y., Yu, E., Li, S., Li, Y., Li, J., Wang, H.: Cityseeker: How do vlms explore embodied urban navigation with implicit human needs? arXiv preprint arXiv:2512.16755  (2025)

\bibitem{wang2025internvl35advancingopensourcemultimodal}
Wang, W., Gao, Z., Gu, L., Pu, H., Cui, L., Wei, X., Liu, Z., Jing, L., Ye, S., Shao, J., Wang, Z., Chen, Z., Zhang, H., Yang, G., Wang, H., Wei, Q., Yin, J., Li, W., Cui, E., Chen, G., Ding, Z., Tian, C., Wu, Z., Xie, J., Li, Z., Yang, B., Duan, Y., Wang, X., Hou, Z., Hao, H., Zhang, T., Li, S., Zhao, X., Duan, H., Deng, N., Fu, B., He, Y., Wang, Y., He, C., Shi, B., He, J., Xiong, Y., Lv, H., Wu, L., Shao, W., Zhang, K., Deng, H., Qi, B., Ge, J., Guo, Q., Zhang, W., Zhang, S., Cao, M., Lin, J., Tang, K., Gao, J., Huang, H., Gu, Y., Lyu, C., Tang, H., Wang, R., Lv, H., Ouyang, W., Wang, L., Dou, M., Zhu, X., Lu, T., Lin, D., Dai, J., Su, W., Zhou, B., Chen, K., Qiao, Y., Wang, W., Luo, G.: Internvl3.5: Advancing open-source multimodal models in versatility, reasoning, and efficiency (2025), \url{https://arxiv.org/abs/2508.18265}

\bibitem{wang2024interactive}
Wang, X., Liu, Y., Song, X., Liu, Y., Zhang, S., Jiang, S.: An interactive navigation method with effect-oriented affordance. In: Proceedings of the IEEE/CVF Conference on Computer Vision and Pattern Recognition. pp. 16446--16456 (2024)

\bibitem{wei2025ground}
Wei, M., Wan, C., Peng, J., Yu, X., Yang, Y., Feng, D., Cai, W., Zhu, C., Wang, T., Pang, J., et~al.: Ground slow, move fast: A dual-system foundation model for generalizable vision-and-language navigation. arXiv preprint arXiv:2512.08186  (2025)

\bibitem{wei2025streamvln}
Wei, M., Wan, C., Yu, X., Wang, T., Yang, Y., Mao, X., Zhu, C., Cai, W., Wang, H., Chen, Y., et~al.: Streamvln: Streaming vision-and-language navigation via slowfast context modeling. arXiv preprint arXiv:2507.05240  (2025)

\bibitem{wijmans2019dd}
Wijmans, E., Kadian, A., Morcos, A., Lee, S., Essa, I., Parikh, D., Savva, M., Batra, D.: Dd-ppo: Learning near-perfect pointgoal navigators from 2.5 billion frames. arXiv preprint arXiv:1911.00357  (2019)

\bibitem{xue2025omninav}
Xue, X., Hu, J., Luo, M., Xie, S., Chen, J., Xie, Z., Quan, K., Guo, W., Xu, M., Chu, Z.: Omninav: A unified framework for prospective exploration and visual-language navigation. arXiv preprint arXiv:2509.25687  (2025)

\bibitem{yokoyama2024vlfm}
Yokoyama, N., Ha, S., Batra, D., Wang, J., Bucher, B.: Vlfm: Vision-language frontier maps for zero-shot semantic navigation. In: 2024 IEEE International Conference on Robotics and Automation (ICRA). pp. 42--48. IEEE (2024)

\bibitem{yokoyama2024hm3d}
Yokoyama, N., Ramrakhya, R., Das, A., Batra, D., Ha, S.: Hm3d-ovon: A dataset and benchmark for open-vocabulary object goal navigation. In: 2024 IEEE/RSJ International Conference on Intelligent Robots and Systems (IROS). pp. 5543--5550. IEEE (2024)

\bibitem{zhang2025embodied}
Zhang, J., Li, A., Qi, Y., Li, M., Liu, J., Wang, S., Liu, H., Zhou, G., Wu, Y., Li, X., et~al.: Embodied navigation foundation model. arXiv preprint arXiv:2509.12129  (2025)

\bibitem{zhang2024uni}
Zhang, J., Wang, K., Wang, S., Li, M., Liu, H., Wei, S., Wang, Z., Zhang, Z., Wang, H.: Uni-navid: A video-based vision-language-action model for unifying embodied navigation tasks. arXiv preprint arXiv:2412.06224  (2024)

\bibitem{zhao2021surprising}
Zhao, X., Agrawal, H., Batra, D., Schwing, A.G.: The surprising effectiveness of visual odometry techniques for embodied pointgoal navigation. In: Proceedings of the IEEE/CVF International Conference on Computer Vision. pp. 16127--16136 (2021)

\bibitem{zhao2026bridgingindooroutdoorgapvisioncentric}
Zhao, Y., Yang, Y., Zhu, Y., Shen, Y., Wang, C., Gu, Z., Shi, P., Guo, W., Xu, M.: Bridging the indoor-outdoor gap: Vision-centric instruction-guided embodied navigation for the last meters (2026), \url{https://arxiv.org/abs/2602.06427}

\bibitem{zheng2024towards}
Zheng, D., Huang, S., Zhao, L., Zhong, Y., Wang, L.: Towards learning a generalist model for embodied navigation. In: Proceedings of the IEEE/CVF Conference on Computer Vision and Pattern Recognition. pp. 13624--13634 (2024)

\end{thebibliography}
\end{document}